\documentclass[journal]{IEEEtran}
\pdfoutput=1
\usepackage{amsfonts}
\usepackage{textcomp}
\usepackage{mathtools}
\usepackage{makecell}
\usepackage{booktabs}
\usepackage{adjustbox}
\usepackage{multicol}
\usepackage{epsf}
\usepackage{algpseudocode}
\usepackage[norelsize, linesnumbered, ruled, lined, boxed, commentsnumbered]{algorithm2e}
\usepackage{qtree}

\newcommand{\nosemic}{\renewcommand{\@endalgocfline}{\relax}}
\newcommand{\dosemic}{\renewcommand{\@endalgocfline}{\algocf@endline}}

\let\oldnl\nl
\newcommand{\nonl}{\renewcommand{\nl}{\let\nl\oldnl}}
\usepackage{relsize}
\usepackage{etoolbox}
\usepackage{arydshln}
\usepackage{lipsum}
\usepackage{enumitem}
\usepackage{rotating}
\usepackage{comment}
\usepackage{url}
\usepackage{float}
\usepackage{booktabs}
\usepackage{setspace}
\usepackage{epstopdf}
\usepackage{adjustbox}  
\usepackage{amsmath}
\usepackage{xcolor}
\usepackage{amssymb}
\usepackage{multirow}
\usepackage{color}
\usepackage{graphicx}
\usepackage{tikz}
\definecolor{DarkBlue}{RGB}{0, 0, 0}
\usetikzlibrary{fit,calc}
\newcommand{\tikzmk}[1]{\tikz[remember picture,overlay] \node (#1) {};\ignorespaces}

\newcommand{\boxitLL}[1]{\tikz[remember picture,overlay]{\node[yshift=3pt,fill=#1,fill opacity=.15,text opacity=1,fit={($(A)+(.005\linewidth,0.13\baselineskip)$)($(B)+(.885\linewidth,.8\baselineskip)$)}] {};}\ignorespaces}

\newcommand{\boxitLLLL}[1]{\tikz[remember picture,overlay]{\node[yshift=3pt,fill=#1,fill opacity=.15,text opacity=1,fit={($(A)+(.005\linewidth,0.13\baselineskip)$)($(B)+(.809\linewidth,.67\baselineskip)$)}] {};}\ignorespaces}
\newcommand{\boxitLLLFULL}[1]{\tikz[remember picture,overlay]{\node[yshift=3pt,fill=#1,fill opacity=.15,text opacity=1,fit={($(A)+(.005\linewidth,0.13\baselineskip)$)($(B)+(.84\linewidth,.69\baselineskip)$)}] {};}\ignorespaces}
\newcommand{\boxitLLLLLLLL}[1]{\tikz[remember picture,overlay]{\node[yshift=3pt,fill=#1,fill opacity=.15,text opacity=1,fit={($(A)+(.005\linewidth,0.13\baselineskip)$)($(B)+(.6845\linewidth,.8\baselineskip)$)}] {};}\ignorespaces}
\SetKwFor{ForAll}{for all}{in parallel do}{end}
\DeclareMathAlphabet{\mathbfit}{OML}{cmm}{b}{it}
\SetKwFunction{Metropolis}{Metropolis}
\SetKwFunction{Neighbor}{Neighbor}
\SetKwFunction{Cost}{Cost}
\SetKwFunction{Sample}{Sample}
\SetKwFunction{Best}{Best}
\SetKwFunction{Move}{Move}
\makeatletter
\def\algbackskip{\hskip-\ALG@thistlm}
\makeatother
\usepackage{microtype}
\usepackage{subfig}

\algnewcommand{\LeftComment}[1]{\Statex \(\triangleright\) #1}
\DeclarePairedDelimiter{\ceil}{\lceil}{\rceil}

\usepackage{caption}
\usepackage[sorting = none, backend = biber, style=numeric-comp,citestyle=ieee]{biblatex} 
\usepackage[utf8]{inputenc}   
\usepackage[T1]{fontenc}
\usepackage{etoolbox}

\usepackage{bm}

\makeatother

\DeclareRobustCommand*{\IEEEauthorrefmark}[1]{%
  \raisebox{0pt}[0pt][0pt]{\textsuperscript{\footnotesize #1}}%
}

\begin{document}
%
\title{Optimization of FPGA-based CNN Accelerators Using Metaheuristics}%
%
%
\author{\IEEEauthorblockN{
Sadiq~M.~Sait\IEEEauthorrefmark{1, 2},
Aiman~El-Maleh\IEEEauthorrefmark{1, 2},
Mohammad~Altakrouri\IEEEauthorrefmark{1}, and
Ahmad~Shawahna\IEEEauthorrefmark{1}
}\\
\IEEEauthorblockA{\IEEEauthorrefmark{1}Department of Computer Engineering, King Fahd University of Petroleum and Minerals, Dhahran-31261, KSA.\\}
\IEEEauthorblockA{\IEEEauthorrefmark{2}Interdisciplinary Research Center for Intelligent Secure Systems, King Fahd University of Petroleum and Minerals, Dhahran-31261, KSA.\\}
\{sadiq, aimane, g201707010, g201206920\}@kfupm.edu.sa
}

\maketitle

\let\thefootnote\relax\footnotetext{This preprint has not undergone peer review (when applicable) or any post-submission improvements or corrections. The Version of Record of this article is published in The Journal of Supercomputing, and is available online at \textbf{\textit{https://doi.org/10.1007/s11227-022-04787-8}}.}

\begin{abstract}
In recent years, convolutional neural networks (CNNs) have demonstrated their ability to solve problems in many fields and  with accuracy that was not possible before. However, this comes with extensive computational requirements, which made general central processing units (CPUs) unable to deliver the desired real-time performance. At the same time, field-programmable gate arrays (FPGAs) have seen a surge in interest for accelerating CNN inference. This is due to their ability to create custom designs with different levels of parallelism. Furthermore, FPGAs provide better performance per watt compared to other computing technologies such as graphics processing units (GPUs). The current trend in FPGA-based CNN accelerators is to implement multiple convolutional layer processors (CLPs), each of which is tailored for a subset of layers. However, the growing complexity of CNN architectures makes optimizing the resources available on the target FPGA device to deliver the optimal performance more challenging. This is because of the exponential increase in the design variables that must be considered when implementing a Multi-CLP accelerator as CNN's complexity increases. In this paper, we present a CNN accelerator and an accompanying automated design methodology that employs metaheuristics for partitioning available FPGA resources to design a Multi-CLP accelerator. Specifically, the proposed design tool adopts simulated annealing (SA) and tabu search (TS) algorithms to find the number of CLPs required and their respective configurations to achieve optimal performance on a given target FPGA device. Here, the focus is on the key specifications and hardware resources, including digital signal processors (DSPs), block random-access memories (BRAMs), and off-chip memory bandwidth. Experimental results and comparisons using four well-known benchmark CNNs are presented demonstrating that the proposed acceleration framework is both encouraging and promising. The SA-/TS-based Multi-CLP achieves 1.31$\times$ $-$ 2.37$\times$ higher throughput than the state-of-the-art Single-/Multi-CLP approaches in accelerating AlexNet, SqueezeNet~1.1, VGGNet, and GoogLeNet architectures on the Xilinx VC707 and VC709 FPGA boards.  
\end{abstract}

\begin{IEEEkeywords}
	Convolutional Neural Network, FPGA, Metaheuristics, Simulated Annealing, Tabu Search, Combinatorial Optimization, NP-Hard Problems.
\end{IEEEkeywords}

%
\IEEEpeerreviewmaketitle


\section{Introduction} \label{sec:introduction}

\IEEEPARstart{C}{onvolutional} neural network (CNN) is a powerful method used for processing data with predefined grid-like topology, such as $1$-D time series in speech recognition~\cite{hu2014mandarin}, $2$-D image retrieval in face detection and recognition~\cite{6903759, farfade2015multi}, and in intelligent transportation systems~\cite{7153896, wang2015pedestrian, 7252029}, to name a few. The CNN proved its effectiveness when it was used to win the ImageNet challenge, an annual competition for visual object recognition, in $2012$, by dropping the classification error record from $26$\% to $15$\%, which was a significant improvement at the time~\cite{krizhevsky2012imagenet}. Since then, more studies and applications have emerged to enhance CNN application in different fields~\cite{8594633, feng2019computer, ghimire2022survey}.

\begin{figure*}[t!]
\centering
\includegraphics[width=0.99\textwidth]{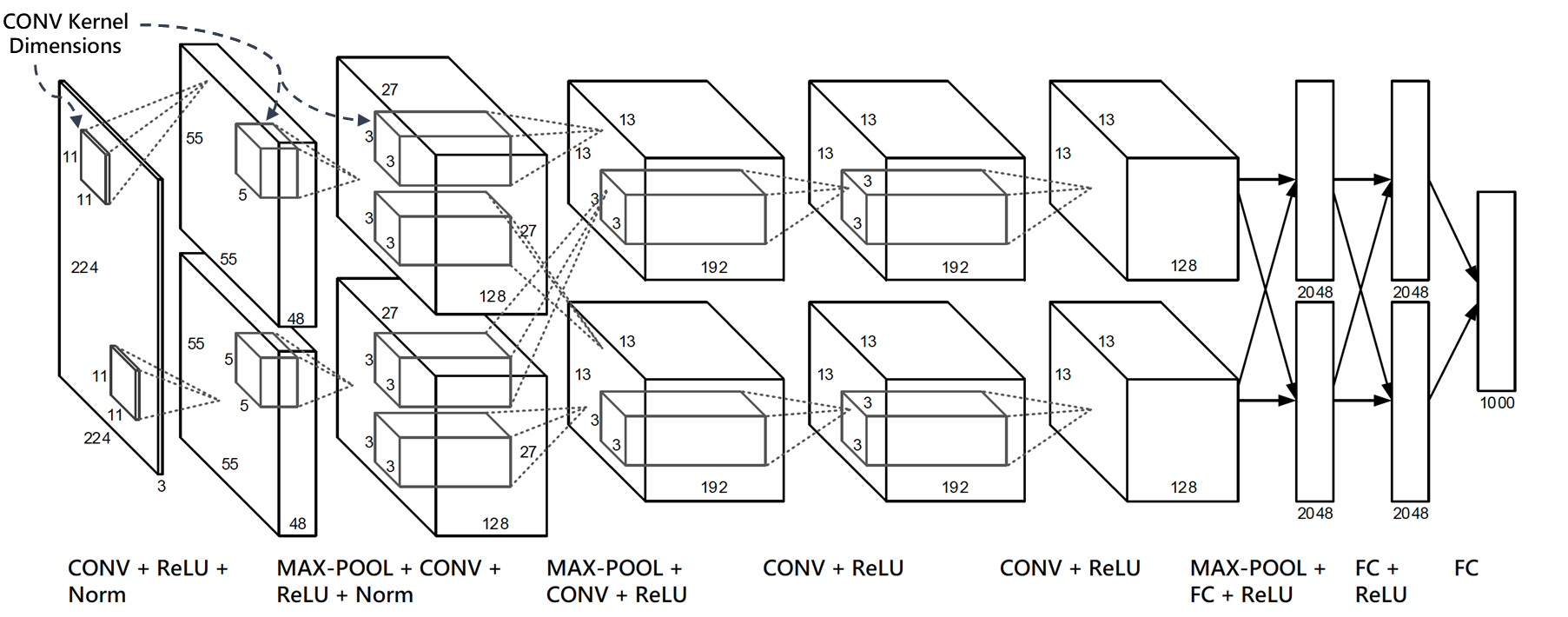}
\caption{An illustration of AlexNet architecture. The input/output feature maps to/from each layer(s) are shown as matrices with their respective dimensions and the operations performed are indicated below.}
\label{fig:alexnet}
\end{figure*}

In general, CNNs consist of an input layer, an output layer, and multiple intermediate hidden layers. The key operations involved in the construction of CNN layers include convolution (CONV), pooling (POOL), and inner product. The CONV layer extracts the unique features from the input image. Specifically, the first CONV layer in the architecture captures low-level features, such as edges, colors, gradients, orientations, etc., while subsequent CONV layers adapt to extract higher-level features, resulting in a wholesome understanding of the processed image.

On the other hand, the POOL layer reduces the dimensionality of the processed data, extracts rotational and positional invariant features, and suppresses noise by applying the $max$, $avg$, or $min$ functions. Last, but not least, the inner product layer, also known as the dense or the fully connected (FC) layer, is a layer whose neurons are connected to all the neurons of its preceding layer, hence the name. It is worth noting that the last FC layer in the architecture is also referred to as the classification layer because it is responsible for decision-making, such as the class score. Here, we must emphasize that there are other layers that can also be used in constructing CNNs such as normalization (Norm) and rectified linear unit (ReLU). The Norm layer is used to smooth the data, whereas the ReLU activation layer helps in learning and modeling complex data.

CNN layers are represented by a set of matrices that reflect the main elements and the relationship between them according to the network structure. Specifically, each layer receives the output matrix of the previous layer as input, which is a set of $2$-D arrays named feature maps (FMs). Then, it performs its operation to produce the FMs for the subsequent layer. Figure~\ref{fig:alexnet} shows a well-known CNN architecture called AlexNet~\cite{krizhevsky2012imagenet}. It consists of $5$ CONV layers each of which is followed by a ReLU activation function, interspersed by $2$ Norm layers, $3$ MAX-POOL layers, and concluded by $3$ FC layers.

\begin{figure}[t!]
\centering
\includegraphics[width=1\linewidth]{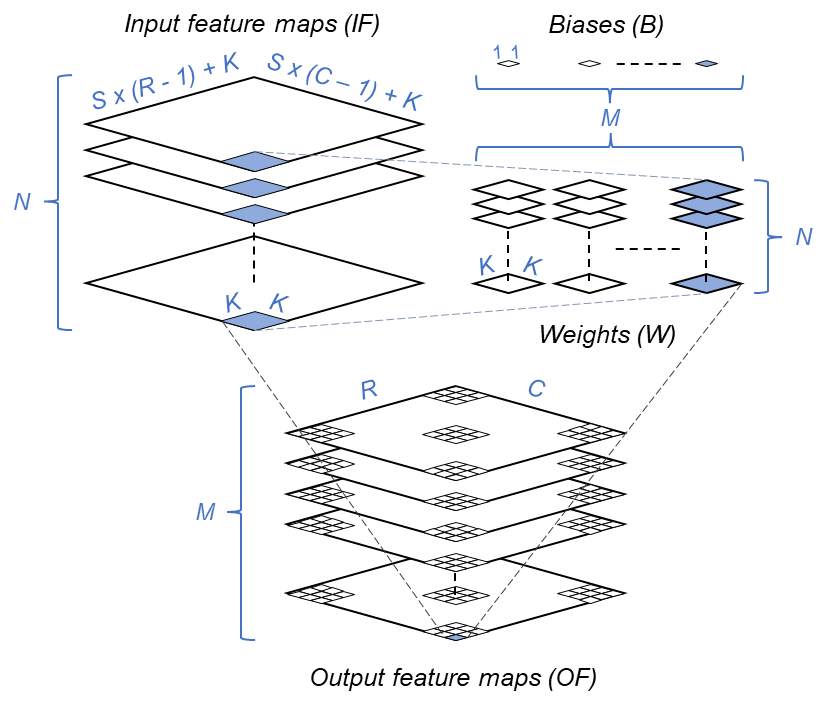}
\caption{An illustration of the convolution layer. The blue feature in the last output feature map is computed by taking the dot-product of the blue weights with the blue highlighted portion of the input feature maps and then adding the blue shaded bias value to the result.}
\label{fig:conv_operation}
\end{figure}

In CNNs, as the name implies, CONV layers are the most critical ones, and their operations constitute over $90$\% of the total computation time~\cite{cong2014minimizing}. The CONV layer receives a set of $N$ FMs from the previous layer. Then, it convolves the input FMs (IF) with a set of $M$ small kernels, also known as filters, using a shifting window that slides over the IF with a stride of size $S$. The kernel is a matrix of numbers called weight parameters whose values are learned during the training phase by the backpropagation algorithm. Each kernel is used to compute one specific output FM, that is, the ${m\textnormal{-th}}$ kernel produces the $m$-th output FM as demonstrated in Figure~\ref{fig:conv_operation}. Thus, the number of output FMs (OF) equals the number of CONV layer kernels. Additionally, each kernel has one bias term (B). Each bias is added to every element in its corresponding output FM to produce the final OF, which form the IF for the next layer.

The number of CONV kernels, their size, and the number of channels are defined by the CNN designers depending on the type of convolution. Specifically, standard convolution and point-wise convolution employ $M$ kernels of size ${K \times K}$ and ${1 \times 1}$, respectively, each of which contains $N$ channels~\cite{howard2017mobilenets}. Conversely, depth-wise convolution uses $N$ single-channel kernels of size ${K \times K}$, where each kernel is applied to one input FM to produce the corresponding output FM, i.e., ${M = N}$. To determine the value of the neuron at position $(r, c)$ of $m$-th output FM, the following computation is performed
\begin{equation}
\begin{aligned}
 \mathbf{OF}\!\left[m\right]\!\left[r\right]\!\left[c\right] = {} \mathbf{B}\!\left[m\right] + & \! \sum_{n = 0}^{N - 1} \sum_{i = 0}^{K - 1} \sum_{j = 0}^{K - 1} \mathbf{W}\!\left[m\right]\!\left[n\right]\!\left[i\right]\!\left[j\right] \times \\
 & \,\,\,\,\, \mathbf{IF}\!\left[n\right]\!\left[S \times r + i\right]\!\left[S \times c + j\right]
\end{aligned}
\label{eq:convolve}
\end{equation}

Here, $\mathbf{B}\!\left[m\right]$ represents the bias parameter of the ${m\textnormal{-th}}$ kernel, ${m \in [M]}$, ${\mathbf{W}\!\left[m\right]\!\left[n\right]\!\left[i\right]\!\left[j\right]}$ denotes the weight parameter at position $(i, j)$ in the $n$-th channel of the $m$-th kernel, ${n \in [N]}$, $i$ and ${j \in [K]}$, ${\mathbf{IF}\!\left[n\right]\!\left[x\right]\!\left[y\right]}$ indicates the neuron at position $(x, y)$ of the $n$-th input FM, and $S$ is the stride size. By applying Equation~\eqref{eq:convolve} to compute all neurons of OF, we can see that the main structure of CONV layer consists of $6$ nested loops as shown in Algorithm~\ref{alg:convolution}. One can also note that these loops contain a massive number of the multiply-accumulate (MAC) operations. Precisely, each layer performs ${M \times R \times C \times N \times K^2}$ MAC operation. Overall, AlexNet requires about ${1.46}$ Giga operations (GOPs) to process a single RGB image of size ${224 \times 224}$ pixels.

The intensive computational requirements of CNNs limit their real-time applications on general central processing units (CPUs). On the other hand, the computational efficiency of graphics processing units (GPUs) and field-programmable gate arrays (FPGAs) make them excellent platforms for accelerating CNNs. However, the critical need for lower power consumption in today's applications, due to CNN deployment on battery-powered devices such as drones and Internet of things (IoT)~\cite{horng2019smart, jiang2021iot}, makes FPGAs more suitable for CNN acceleration~\cite{8594633}. This is because of the high power efficiency, also known as performance per watt, that FPGAs offer.

With regards to model parameters, each standard CONV layer learns ${M \times N \times K^2}$ weight parameters. For example, AlexNet's first CONV layer receives $3$ input FMs, uses $11 \times 11$ kernels, and produces $96$ output FMs, resulting in $34,848$ parameters. Overall, AlexNet has over $60$ million parameters which need about $240$~MB of memory space. The memory requirements for FMs and model parameters exceed what commercially available FPGAs can provide in on-chip memory. Thus, they must be stored in off-chip memory and transferred to on-chip memory during computation on need basis. The significant amount of storage and external memory bandwidth required become a performance bottleneck. Therefore, many attempts have been made to maximize the throughput of CNN applications by designing a hardware accelerator known as a convolutional layer processor (CLP)~\cite{li2016high, zhang2015optimizing, suda2016throughput, shen2017maximizing, Lu2017FlexFlow}.

\begin{algorithm}[!t]
\small
\SetAlgoLined
\DontPrintSemicolon
\KwIn{${\textnormal{The input feature maps }(\mathbf{IF})\textnormal{, weight }(\mathbf{W})\textnormal{ and bias}}$ ${(\mathbf{B})\textnormal{ parameters, number of input feature maps }(N)}$ ${\textnormal{and output feature maps }(M)\textnormal{, size of each output}}$ ${\textnormal{feature map, rows }(R)\textnormal{ and columns }(C)\textnormal{, kernel size}}$ ${(K)\textnormal{, and}}$ ${\textnormal{window stride size }(S)}$.} 
\KwOut{The output feature maps ($\mathbf{OF}$).}
\SetKwFunction{CONV}{CONV}
\SetKwProg{Fn}{Procedure}{:}{}
\nonl\SetAlgoNoLine\Fn{\CONV{$\mathbf{IF}, \mathbf{W}, \mathbf{B}, N, M, R, C, K, S$}}{
\SetAlgoLined
\tikzmk{A}{\For{$m \gets 0$ \KwTo $M - 1$}{
    \For{$r \gets 0$ \KwTo $R - 1$}{
        \For{$c \gets 0$ \KwTo $C - 1$}{
            $\mathbf{OF}\!\left[m\right]\!\left[r\right]\!\left[c\right] \gets \mathbf{B}\!\left[m\right]$
            
            \For{$n \gets 0$ \KwTo $N - 1$}{
                \For{$i \gets 0$ \KwTo $K - 1$}{
                    \For{$j \gets 0$ \KwTo $K - 1$}{
                        ${\mathcal{P} \gets \mathbf{W}\!\left[m\right]\!\left[n\right]\!\left[i\right]\!\left[j\right] \times}$ ${\,\,\,\,\,\,\,\,\,\,\:\mathbf{IF}\!\left[n\right]\!\left[S \times r + i\right]\!\left[S \times c + j\right]}$
                        
                        ${\mathbf{OF}\!\left[m\right]\!\left[r\right]\!\left[c\right] \gets \mathbf{OF}\!\left[m\right]\!\left[r\right]\!\left[c\right] + \mathcal{P}}$
                    }
                }
            }
        }            
    }
}

}\tikzmk{B}\boxitLL{yellow}
\KwRet $\;\mathbf{OF}$

}
\caption{Convolution Algorithm.}
\label{alg:convolution}
\end{algorithm}

The CLP optimizes the implementation of CONV layers by applying loop transformations to Algorithm~\ref{alg:convolution}. Specifically, the loop unrolling technique is used to maximize the parallelism of CONV computations by replicating the hardware resources of MAC unit. Furthermore, loop tilling and local memory promotion techniques are adopted to reduce off-chip memory accesses and maximize data sharing and reuse. Here, we must emphasize that the CLP design is parametrized by the number of input/output FMs to/from CONV layers as well as their dimensions. However, different CONV layers, even those in the same architecture, vary considerably in their configurations ($N$, $M$, $R$, $C$, $K$, and $S$).

To overcome this problem, three main design schemes have been proposed in the literature. In the first approach, such as in~\cite{li2016high}, a CLP is modeled for each CONV layer. In this way, a CNN of $L$ CONV layers is accelerated using $L$ CLPs that process $L$ independent images in a pipelined fashion. Even though this approach optimizes the computations of each CONV layer, it suffers from non-negligible latency and bandwidth overheads. This is due to the need for orchestrating the off-chip memory accesses for a large number of CLPs. Additionally, dividing the limited on-chip memory among many CLPs reduces the overall data locality. Last, but not least, implementing a dedicated controller for each CLP leaves them with insufficient resources for computation.

\begin{figure}[t!]
\centering
\includegraphics[width=\linewidth]{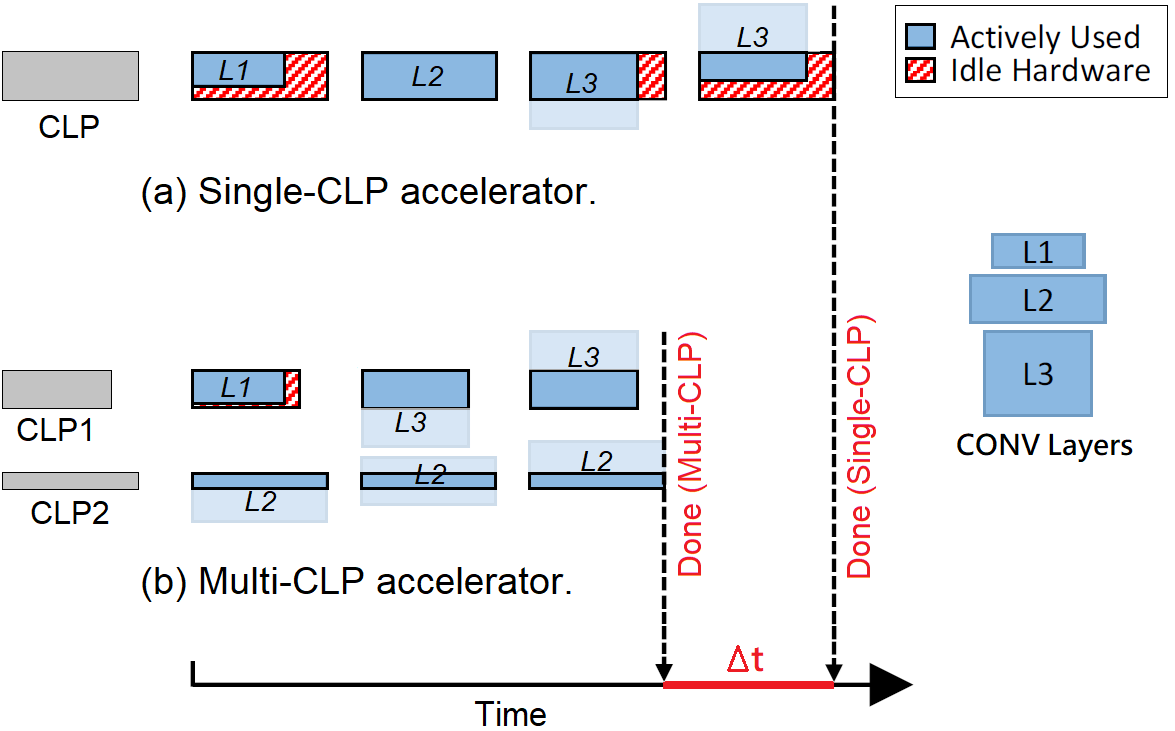}
\caption{The operation of (a) Single-CLP accelerator and (b)~$\textnormal{Multi-CLP}$ accelerator on CNN with three CONV layers. The dimensions of the hardware ($CLP$, $CLP1$, and $CLP2$) and the CONV layers ($L1$, $L2$, and $L3$) are represented by the size and shape of the boxes.}
\label{fig:single_CLP_vs_multi_CLPs}
\end{figure}

The second approach, on the other hand, designs a single, unified CLP based on the optimal parameters that achieve the lowest overall latency~\cite{zhang2015optimizing, suda2016throughput}. The globally optimized CLP is then used to iteratively process CONV layers, one layer at a time. However, a CLP that provides the best performance across all layers is not necessarily optimal in utilizing its hardware resources. This is due to the radically varying CONV configurations. Considering the three CONV layers shown in Figure~\ref{fig:single_CLP_vs_multi_CLPs} as an example, one can note that the Single-CLP accelerator in Figure~\ref{fig:single_CLP_vs_multi_CLPs}a is underutilized while processing $L1$ and the portions of $L3$.

To alleviate this issue, the third approach implements multiple CLPs and optimally distributes the hardware resources among them~\cite{shen2017maximizing, Lu2017FlexFlow}. In Figure~\ref{fig:single_CLP_vs_multi_CLPs}b, the same hardware resources used to implement the $\textnormal{Single-CLP}$ accelerator in Figure~\ref{fig:single_CLP_vs_multi_CLPs}a are partitioned into two CLPs ($CLP1$ and $CLP2$). Thereafter, by mapping $L1$ and $L3$ to $CLP1$ and $L2$ to $CLP2$, the $\textnormal{Multi-CLP}$ design achieves $\Delta t$ reduction in the overall execution time compared to the $\textnormal{Single-CLP}$ design.

The problem now is, how to determine the appropriate number of CLPs to use, on what basis to assign a CONV layer to a CLP, how many hardware resources each CLP can utilize, and how to determine the appropriate tiling factors to assign to each CONV layer? This intractable design space makes finding near-optimal configurations by exhaustive search algorithms almost impossible. Thus, intelligent techniques are needed to efficiently explore the design space for the optimal configurations of $\textnormal{Multi-CLP}$ design that improve the throughput of CNN applications.

Metaheuristics algorithms have proved their capabilities for finding optimal solutions for several NP-hard problems, with very efficient performance~\cite{osman1996meta}. However, very few studies exist in the current literature that have proposed and exploited the use of metaheuristics algorithms in optimizing CNNs. Specifically, the standard genetic algorithm was employed in~\cite{suda2016throughput} to explore the design space for an optimal Single-CLP accelerator. On the other hand, the authors in~\cite{rere2015simulated} adopted the simulated annealing algorithm to improve the training of the LeNet-5 architecture.

In this paper, we present a systematic methodology for the optimization of the throughput of an FPGA-based accelerator with $\textnormal{Multi-CLP}$ implementation. Specifically, we present an analytical and empirical design scheme for accurate modeling of cost, in terms of hardware resources used, and performance for a $\textnormal{Multi-CLP}$ design given its parameters. Then, we employ simulated annealing and tabu search metaheuristic algorithms to search for the optimal $\textnormal{Multi-CLP}$ design, constrained by the given CNN architecture and target FPGA specifications. As a case study, we implemented CNN accelerators for four well-known benchmark architectures, namely, AlexNet~\cite{krizhevsky2012imagenet}, SqueezeNet~$1.1$~\cite{iandola2016squeezenet}, VGGNet~\cite{simonyan2014very}, and GoogLeNet~\cite{szegedy2015going} on VC707 and VC709 FPGA boards and compared them with previous approaches. Our implementation achieves a performance of $113.92$ Giga floating-point operations (GFLOPs) under $100$ MHz working frequency. The key contributions of this work are summarized as follows:
\begin{itemize}
\item Designing a CNN accelerator with loop tiling, loop unrolling, and loop pipelining techniques. 
 \item Modeling CNN implementation and identifying the critical optimization variables for multiple CLPs design on FPGA platforms.
 \item Clarifying the complexity of finding the optimal configurations for multiple CLPs accelerator, i.e., defining the problem statement and the resultant intractable design space.
 \item Introducing a metric to quantify the dimensional mismatches between CLP dimensions and CONV layer dimensions, and use it to improve CNN throughput.
 \item Presenting computational performance and resource usage models for cost estimation of a candidate multiple CLPs design.
 \item Proposing a metaheuristic-based optimization framework that employs the estimation models to efficiently explore the design space using simulated annealing and tabu search algorithms, and find the optimal multiple CLPs design for CNNs on FPGAs based on performance and resources constraints.
 \item Validating the solutions obtained and comparing them with related works.
\end{itemize}

The rest of the paper is organized as follows. Section~\ref{sec:related_work} reviews previous work related to accelerating CNNs using FPGAs. In Section~\ref{sec:cnn_accelerator_design}, we discuss the optimization techniques applied to achieve efficient implementation of CNNs on FPGA platforms. Furthermore, we present an analytical and empirical model to accurately predict the performance and hardware resources required for a given design variant. Section~\ref{sec:design_space_exploration} presents the methodology used for exploring the design space to achieve an efficient implementation of $\textnormal{Multi-CLP}$ hardware accelerator for CONV layers. The results of our experiments are presented and discussed in Section~\ref{sec:results}, and the paper is concluded in Section~\ref{sec:conclusion}.

\section{Related Work} \label{sec:related_work}

In this section, we review the existing approaches targeting the optimization of CONV implementation on FPGA platforms. Although CNN structures perform well for their intended applications, they have the potential to be further optimized with minimal impact on accuracy. For example, the values of FMs and weight parameters are originally represented as $32$-bit floating-point numbers. However, it has been demonstrated that fewer bits can be used to represent these values without a noticeable accuracy drop~\cite{shawahna2022fxp}. This reduces the hardware requirements for CNN implementation as well as reduces the inference latency.

The authors in~\cite{cho2021fpga} proposed an accelerator for LeNet-5 architecture to perform handwritten digits classification. The proposed strategy is based on three major aspects; loop parallelization to utilize resources, fixed-point data optimization to find the minimum number of bits that maintains accuracy level, and finally implementing MAC approximate units through logic blocks such as look-up tables (LUTs) and flip-flops (FFs) rather than using high-precision digital signal processors (DSPs). With these optimizations, the authors achieve less memory usage and reduced network latency. However, the solution is problem specific and deals with a relatively small CNN. Large CNNs would require advanced techniques to achieve the full potential of FPGA resources.

In~\cite{zhang2015optimizing}, the authors adopted the polyhedral-based data dependence analysis~\cite{pouchet2013polyhedral} to optimize the computations and memory access operations in CONV layers. Specifically, they proposed an analytical design scheme to estimate the computational performance of a given CONV design. The loop transformations are employed to derive all possible CONV designs. The goal behind employing these techniques is to fully utilize the hardware resources provided by the target FPGA platform for effective acceleration. After enumerating all candidate solutions, the roofline performance model~\cite{williams2009roofline} is used to identify the optimal design for each layer.

Here, we must emphasize that different CONV layers have different structures, and therefore varying optimal loop unrolling and loop tiling factors. Considering AlexNet architecture discussed in Figure~\ref{fig:alexnet}, the optimal unrolling factors $\langle T_{n}, T_{m} \rangle$ for the second and third CONV layers are $\langle 24, 20 \rangle$ and $\langle 5, 96 \rangle$, respectively, where $T_{n}$ and $T_{m}$ are the unrolling factors for the input FMs and output FMs, respectively. Hence, designing a CLP to accelerate these differently structured layers requires complex hardware implementation to reconfigure the computational units and their interconnects.

To overcome this issue, the authors in~\cite{zhang2015optimizing} designed the CLP based on the uniform cross-layer unrolling factors. More precisely, they found the unified unrolling factors $\langle T^{'}_{n}, T^{'}_{m} \rangle$ that maximize the overall performance. Accordingly, the computational engine of the CLP is implemented as $T^{'}_{m}$ duplicated tree-shaped poly structures. These structures receive identical $T^{'}_{n}$ values from the input FMs and each structure receives $T^{'}_{n}$ weights from one of the $T^{'}_{m}$ kernels. Each structure multiplies its data using $T^{'}_{n}$ digital multipliers. Then, an adder tree is used to accumulate the multipliers outputs as well as the previous partial result. Furthermore, they used the double-buffering technique to overlap the off-chip memory and CLP buffers data transfers with computation.

Suda et al.~\cite{suda2016throughput} proposed an OpenCL-based framework for accelerating CNN inference. Focusing on CONV layers, the authors reformulated the CONV operation into a matrix multiplication operation. Specifically, they organized CONV kernels as a $2$-D matrix of size $M \times (N \times K \times K)$. Similarly, the input FMs were flattened and rearranged as a $2$-D matrix of $(N \times K \times K)$ rows and $(R \times C)$ columns. In this way, the output of CONV layer is calculated by multiplying these two matrices. 

To speed up CNN operations, the authors followed the same strategy in~\cite{zhang2015optimizing} and modeled each type of CNN layer using unified loop unrolling factors. Precisely, they unrolled the output matrix of CONV operation in both dimensions by the factor $T_{out}$. Thus, they designed a CLP consisting of $T_{out} \times T_{out}$ structures, each of which computes an output feature. Accordingly, kernel weights and input FMs matrices are tiled into blocks of size $T_{out} \times T_{out}$. On each iteration, a tile from the kernel weight matrix and a tile from the input FMs matrix are fetched into the on-chip memory. Then, each CLP structure performs $T_{out}$ MAC operations on a row of kernel weights tile and a column of input FMs tile.

To further improve the throughput of CONV layer, the computations on the inputs to CLP structures were also unrolled using the factor $T_{in}$ so as to perform $T_{in}$ MAC operations, out of the $T_{out}$, in parallel.
To find the optimal unrolling configuration, the authors model the execution time of each layer as a function of the unrolling factors. Then, the standard genetic algorithm was used to explore the design space for the minimum overall execution time considering FPGA resource constraints. Hence, this approach is referred to as $\textnormal{GA-based}$ $\textnormal{Single-CLP}$. It is noteworthy that the high-level synthesis tool employed to compile their OpenCL codes to hardware restricted $T_{in}$ to be from the set $\{1, 2, 4, 8, 16\}$ and $T_{out}$ to be integer multiplicative of $T_{in}$.

The FPGA-based CNN accelerators discussed earlier employed a single globally-optimized CLP design to maximize the overall throughput. However, using a CNN accelerator with uniform unrolling factors leads to sub-optimal performance for some CONV layers due to their significantly varying dimensions, which affects the overall performance. For example, the cross-layer optimization for AlexNet layers in~\cite{zhang2015optimizing} increases the total execution cycles of the layer-based optimization by $44,442$ cycles. Moreover, following the methodology in~\cite{zhang2015optimizing} to derive the optimal $\textnormal{Single-CLP}$ design for SqueezeNet~$1.1$ implementation on Virtex-7 690T FPGA, the results show that the dynamic utilization of CLP’s MAC units is less than $77$\%~\cite{shen2017maximizing}.

To improve the throughput of the $\textnormal{Single-CLP}$ design, Shen et al.~\cite{shen2017maximizing} proposed to partition the available hardware resources between multiple CLPs. In doing so, they introduced a two-step iterative algorithm that searches for candidate partitions in the first step and then optimizes the tiling factors of each CONV layer in the second step. Specifically, the algorithm starts with a predefined target performance aiming to find a design with such a performance. 

During the first phase, it generates several candidate partitions of computational resources. For each candidate design, the number of CLPs is set to the partition size. Then, the methodology in~\cite{zhang2015optimizing} is adopted to compute $\langle T_n, T_m \rangle$ for each CLP. The last process in this phase distributes CNN layers on the adopted CLPs. To facilitate this process, the authors proposed to order memory-bounded layers based on their need for computation and communication, whereas computational-bounded layers are ordered based on the difference between the number of input FMs and output FMs.

 With such ordering, the authors assume that similar behavioral layers will be neighboring. Thus, they constrain the algorithm to only assign neighboring layers to the same CLP. At the end of the first phase, the performance of candidate designs is evaluated. If no one meets the target performance, the algorithm slightly reduces the target performance and repeats the first phase. When a valid design is found, the algorithm moves on to the second phase and computes $\langle T_r, T_c\rangle$ based on the methodology in~\cite{zhang2015optimizing}. The parameters for the design with the minimum bandwidth requirement is then used to configure a generic CLP template in high-level synthesis. 

In this work, we improve the previously discussed works by efficiently partitioning hardware resources between multiple CLPs. In doing so, we employ intelligent metaheuristics to help explore the intractable design space for multiple CLPs design that is more optimized than that achieved with conventional iterative algorithms. This results in better utilization of FPGA compute resources, improving CNN throughput. 

Furthermore, unlike the work in~\cite{suda2016throughput} which constraints the unrolling factors to be from a predefined set, this work allows the unrolling factors to be whatever value that yields the highest performance. Last, but not least, the authors in~\cite{shen2017maximizing} have specified the number of CLPs in multiple CLPs design to be $6$ at most. This is because they want to keep the optimization time of their algorithm within an acceptable amount of time. This paper proposes a metaheuristic-based optimization framework that is efficient in exploring design space and takes less than a minute to provide an optimal $\textnormal{Multi-CLP}$ design on a single CPU.

\section{CNN Accelerator Design on FPGA Platforms} \label{sec:cnn_accelerator_design}

In this section, we discuss the optimization techniques applied to achieve efficient implementation of CNNs on FPGA platforms. A hardware accelerator referred to as CLP is typically designed to improve CONV layer throughput. Here, the focus is on CONV layers because they are the most computationally intensive layers. Furthermore, we present an analytical and empirical model to accurately predict the performance and hardware resources required for a given design variant.

\subsection{Optimizing CLP Computation and Memory Access} \label{sec:optimizing_CLP_computation_and_memory_access}

The computation of the CONV layer on FPGAs requires a memory space to store input FMs, output FMs, and all kernel weights. For instance, the second CONV layer in the VGGNet architecture demands $12.25$~MB, $0.14$~MB, and $12.25$~MB for input FMs, kernel weights, and output FMs, respectively. However, FPGA platforms usually contain limited on-chip memory. For example, Xilinx Virtex-7 485T FPGA can store a maximum of $4.52$~MB of data on-chip. Thus, we employ the loop tiling technique to overcome this issue. Initially, all the data required for calculation is stored in the external memory. Then, the CONV operation is iteratively performed on a small portion of data, called a tile or block. Each tile is loaded and cached in on-chip buffers before being fed into the computational engine of the CLP.

However, loop tiling opens several design challenges that affect CLP performance. Improper tiling may degrade the efficiency of data reuse and the parallelism of data processing. Therefore, deciding whether to tile a loop and with what factor plays an important role in the performance that can be achieved. Typically, CONV kernels are small in size, ${K \leq 11}$. Hence, the loops over the kernel dimensions, the loop iterators $i$ and $j$ in Algorithm~\ref{alg:convolution}, are not tiled.

On the other hand, the loops over the $N$ input FMs, the $M$ output FMs, and the dimensions, rows ($R$) and columns ($C$), of each output FM are tilted with the factors $T_n$, $T_m$, $T_r$, and $T_c$, respectively. In other words, each of these loops is transformed into two loops; an outer loop that iterates over the tiles, and an inner loop that iterates over the elements in each tile. Note that we denote the loop iterators over the tile elements of $N$, $M$, $R$, and $C$ as $nt$, $mt$, $rt$, and $ct$, respectively.

Additionally, the CONV algorithm is a good candidate for parallelism. Thus, we adopt the loop unrolling technique to speed up CONV operations. Specifically, loop unrolling utilizes the available FPGA computation resources to maximize the parallelism of the MAC units. Even though loop unrolling considerably improves CLP throughput, it imposes complex connection topologies and affects CLP operating frequency. To mitigate these shortcomings, we only unroll input FMs and output FMs loops based on their tiling factors.

More precisely, the loops indicated by the iterators $nt$ and $mt$ are unrolled with the factors $T_n$ and $T_m$, respectively. In this way, all their operations are performed in parallel. To avoid loop-carried dependence, we transformed unrolled loops to the innermost level. Moreover, the operations of the unrolled loops are fully-pipelined to improve system throughput. The CONV algorithm after optimization for loop tiling, loop unrolling, and loop pipelining is demonstrated in Algorithm~\ref{alg:convolution_unrol_tile}. Note that the nested loops are reordered to maximize data reuse, thus reducing off-chip memory accesses. Determining the optimal value of tiling factors for each CNN architecture will be discussed later in Section~\ref{sec:design_space_exploration}.

\begin{figure}[t!]
\centering
\includegraphics[width=0.49\textwidth]{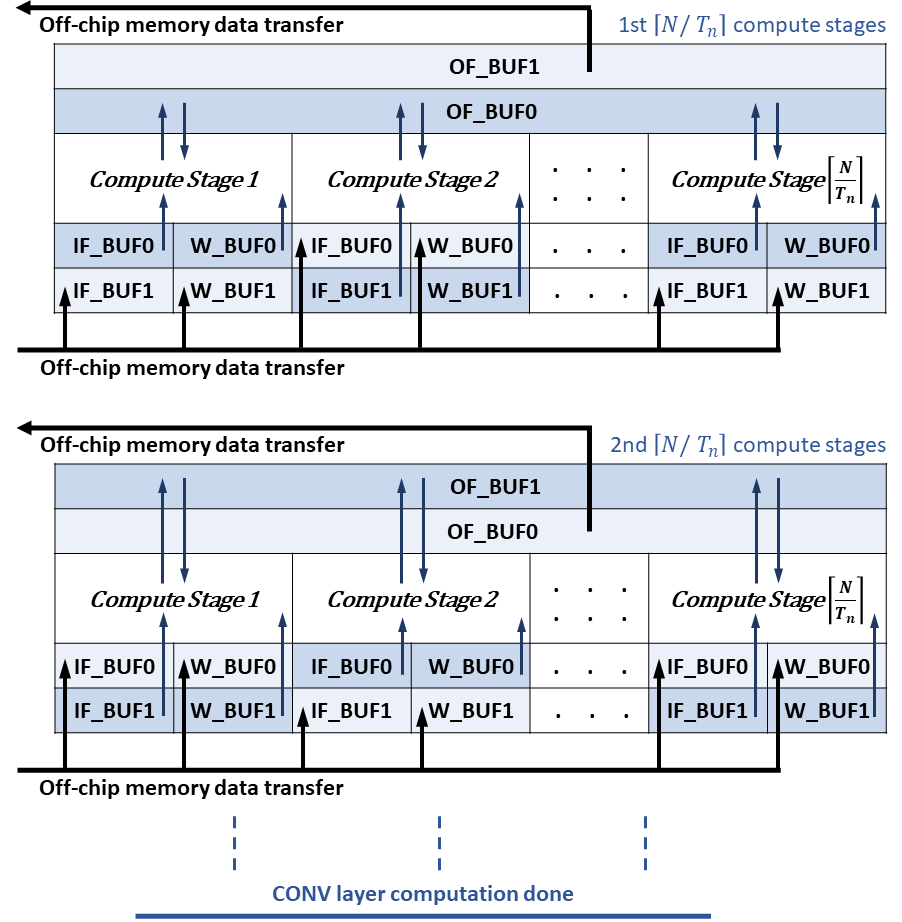}
\caption{CLP computation and ping-pong buffer structure.}
\label{fig:CLP_ping_pong_buffer_structure}
\end{figure} 

\begin{algorithm*}[!t]
\small
\SetAlgoLined
\DontPrintSemicolon
\KwIn{${\textnormal{The input feature maps }(\mathbf{IF})\textnormal{, weight }(\mathbf{W})\textnormal{ and bias }(\mathbf{B})\textnormal{ parameters, number of input feature maps }(N)\textnormal{ and their tiling}}$  ${\textnormal{factor }(T_n)\textnormal{, number of output feature maps }(M)\textnormal{ and their tiling factor }(T_m)\textnormal{, size of each output feature map, rows }(R)}$ ${\textnormal{and columns }(C)\textnormal{, and their corresponding tiling factors, }(T_r)\textnormal{ and }(T_c)\textnormal{, kernel size }(K)\textnormal{, and}}$ ${\textnormal{window stride size }(S)}$.}

\KwOut{The output feature maps ($\mathbf{OF}$).}
\SetKw{KwBy}{by}
\SetKwFunction{CONV}{CONV}
\SetKwProg{Fn}{Procedure}{:}{}
\nonl\SetAlgoNoLine\Fn{\CONV{$\mathbf{IF}, \mathbf{W}, \mathbf{B}, N, T_n, M, T_m, R, T_r, C, T_c, K, S$}}{
\SetAlgoLined
\For{$r \gets 0$ \KwTo $R - 1$ \KwBy $T_r$}{
    \For{$c \gets 0$ \KwTo $C - 1$ \KwBy $T_c$}{
        \For{$m \gets 0$ \KwTo $M - 1$ \KwBy $T_m$}{${\mathbf{OF \_ BUF} \gets \mathrm{Broadcast}\!\left(\mathbf{B}\!\left[m : m + T_m\right], T_r, T_c\right)}$
            
            \For{$n \gets 0$ \KwTo $N - 1$ \KwBy $T_n$}{
                \tikzmk{A}{${\mathbf{IF \_ BUF} \gets \mathbf{IF}\!\left[n : n + T_n\right]\!\left[S \times r : S \times \left(r + T_r - 1\right) + K\right]\!\left[S \times c : S \times \left(c + T_c - 1\right) + K\right]}$
                
                 ${\mathbf{W \_ BUF} \gets \mathbf{W}\!\left[m : m + T_m\right]\!\left[n : n + T_n\right]\!\left[0 : K\right]\!\left[0 : K\right]}$
                 
                 }\tikzmk{B}\boxitLLLL{red}
                \For{$i \gets 0$ \KwTo $K - 1$}{
                    \For{$j \gets 0$ \KwTo $K - 1$}{
                        \For{$rt \gets 0$ \KwTo $\min(T_r - 1, R - r - 1)$}{
                            \For{$ct \gets 0$ \KwTo $\min(T_c - 1, C - c - 1)$}{
                                \tikzmk{A}{\ForAll(\textcolor{DarkBlue}{\Comment{\text{Unroll}}}){$mt \gets 0$ \KwTo $T_m - 1$}{
                                    \ForAll(\textcolor{DarkBlue}{\Comment{\text{Unroll}}}){$nt \gets 0$ \KwTo $T_n - 1$}{
                                        ${\mathcal{P} \gets \mathbf{IF \_ BUF}\!\left[nt\right]\!\left[S \times rt + i\right]\!\left[S \times ct + j\right] \times \mathbf{W \_ BUF}\!\left[mt\right]\!\left[nt\right]\!\left[i\right]\!\left[j\right]}$
                                        
                                        ${\mathbf{OF \_ BUF}\!\left[mt\right]\!\left[rt\right]\!\left[ct\right] \gets \mathbf{OF \_ BUF}\!\left[mt\right]\!\left[rt\right]\!\left[ct\right] + \mathcal{P}}$
                                    }
                                }
                                }\nonl\tikzmk{B}\vskip-10pt\vskip-10pt\rlap{\boxitLLLLLLLL{yellow}}
                            }
                        }            
                    }
                }
            }
            \tikzmk{A}{${\mathbf{OF}\!\left[m : m + T_m\right]\!\left[r : r + T_r\right]\!\left[c : c + T_c\right] \gets \mathbf{OF \_ BUF}}$
            
            }\nonl\tikzmk{B}\vskip-10pt\vskip-10pt\rlap{\boxitLLLFULL{blue}}
        }
    }
}

\KwRet $\;\mathbf{OF}$

}
\caption{Optimized Convolution Algorithm.}
\label{alg:convolution_unrol_tile}
\end{algorithm*}

The optimized CONV algorithm can be divided into two stages; the memory data transfer stage, and the data computation stage. In the data transfer phase, highlighted in red and blue, the input FMs on-chip buffer (IF\_BUF) and weight parameters on-chip buffer (W\_BUF) are filled with a block of input FMs and a block of kernel weights, respectively. Later, the output FM data block contained in the output FMs on-chip buffer (OF\_BUF) is copied back to external memory as shown in line $23$. Note that loop tiling factors control the size of these buffers, and thus the amount of data that is moved for each buffer refill or write-out.

On the other hand, the operations of the computation phase, shaded in yellow, are unrolled based on $T_n$ and $T_m$ tiling factors. Thus, loop tiling factors also control how CLP computational engine is constructed. Specifically, the computational engine is modeled as $T_{m}$ duplicated MAC tree tiles. Each of which receives $T_{n}$ different weights but they all share the same $T_{n}$ input features. Each MAC tree tile multiplies its data using $T_{n}$ digital multipliers. Finally, an adder tree is used to accumulate the output of the multipliers with the previous partial result.

To alleviate, or even prevent, blocking of CLP computations due to external memory data transfer, CLP adopts the double-buffering scheme for all on-chip buffers. In other words, each on-chip buffer can be logically considered as two independent sets operating in a ping-pong manner as illustrated in Figure~\ref{fig:CLP_ping_pong_buffer_structure}. In the first compute stage, the computational engine processes the features and weights from the input buffer set $0$ (IF\_BUF$0$ and W\_BUF$0$). During the same time, the features and weights required for the second compute stage are copied from external memory to input buffer set $1$ (IF\_BUF$1$ and W\_BUF$1$).

The next compute stage does the same but uses the opposite input buffers. That is why it is called the ping-pong buffer structure. With regards to the resulting output FMs, the output features from the first $\ceil*{N \, / \, T_n}$ compute stages are stored in the output buffer set $0$ (OF\_BUF$0$). During these stages, the content of the OF\_BUF$1$, which are the results of the previous $\ceil*{N \, / \, T_n}$ compute stages, is transferred to off-chip memory. Note that the CLP repeats this entire process several times to cover all the computations of CONV layer. As evident from Figure~\ref{fig:CLP_ping_pong_buffer_structure}, double-buffering causes data transfer time to overlap with computation. 

\subsection{Multi-CLP Accelerator Design} \label{sec:multi_clp_accelerator_design}

In the previous section, we discussed an optimized CLP design that processes CONV layers of a CNN architecture iteratively one after another. Given that CNN layers have radically varying dimensions, designing a single CLP to process all layers may be ideal for a particular CONV layer, or some CONV layers, but causes under-utilization of computational resources in others, affecting overall performance. This is because of the mismatch between the CLP dimensions ($T_n$ and $T_m$) and the CONV layers dimensions ($N$ and $M$).

\begin{figure*}[t!]
\vskip 5pt
\centering
\includegraphics[width=1.0\textwidth]{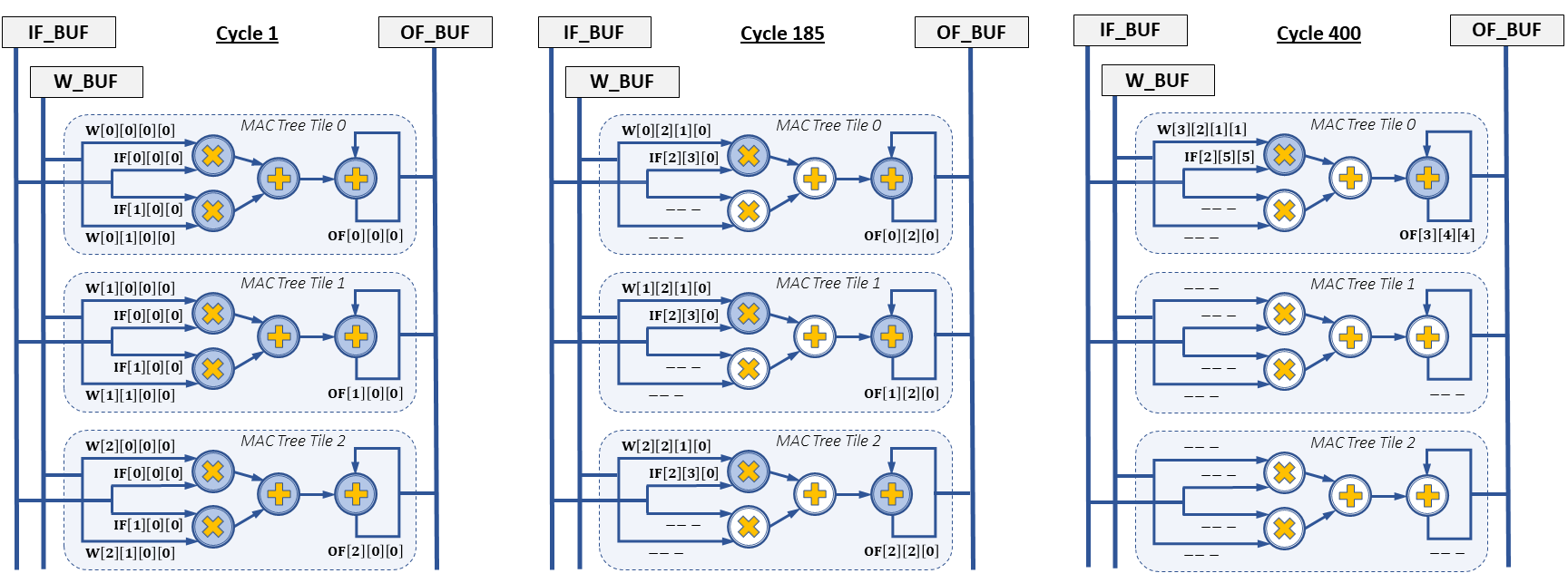}
\caption{An example of a Single-CLP design with $\langle T_{n}, T_{m}, T_{r}, T_{c} \rangle$ factors set to $\langle 2, 3, 2, 2 \rangle$. The CLP processes a CONV layer with $\langle N, M, R, C, K, S \rangle$ configurations equal to $\langle 3, 4, 5, 5, 2, 1 \rangle$.}
\label{fig:single_CLP_example}
\end{figure*}

Considering AlexNet layers illustrated in Figure~\ref{fig:alexnet} as a motivating example, we found that, on average, $78$\% of the computational resources were idle, or doing useless work, while processing the first CONV layer. Note that we followed the methodology in~\cite{zhang2015optimizing} to find the cross-layer optimized $T_n$ and $T_m$. The utilization of ${g\textnormal{-th}}$ CLP computational resources while it is processing CONV layer $\ell$ can be quantified by
\begin{equation}
\label{eq:CONV_utilization}
\lambda_\ell^{(g)} = \frac{\left(N_\ell \, / \, T_n^{(g)}\right) \times \left(M_\ell \, / \, T_m^{(g)}\right)}{\ceil*{N_\ell \, / \, T_n^{(g)}} \times \ceil*{M_\ell \, / \, T_m^{(g)}}}
\end{equation}

To illustrate the computation resource utilization problem, Figure~\ref{fig:single_CLP_example} shows a simple example of a CLP designed with factors $\langle T_n, T_m, T_r, T_c \rangle$ equal to $\langle 2, 3, 2, 2 \rangle$. The CLP is used to process a CONV layer whose configurations $\langle N, M, R, C, K, S \rangle$ are $\langle 3, 4, 5, 5, 2, 1 \rangle$. One can notice that $N$ and $M$ are not a perfect multiple of $T_n$ and $T_m$, respectively. In particular, $T_n$ needs two iterations to cover $N$. The first iteration covers the first two input FMs, leaving one input FM for the second iteration. The same goes for $T_m$ where one output FM is left for the second iteration.

During the first cycle, the CLP works on complete tiles, therefore all of the $T_m$ MAC tree tiles are in full use. On the other hand, the CLP uses a partially filled tile of input FMs in cycle $185$, during which only $T_m$ (out of $T_n \times T_m$) multipliers and adders do useful calculations. To make matters worse, during the last cycle, cycle $400$, which computes the output feature OF[$3$][$4$][$4$], neither IF\_BUF nor W\_BUF is completely full. Thus, only half of the resources of the first MAC tree tile are used. The combined effect of $T_n $ and $T_m$ dimensional mismatches leads to utilizing only $50$\% of the computational resources based on Equation~\eqref{eq:CONV_utilization}.

To overcome the issue of under-utilization, and thus improve CNN throughput, the CLP can be designed with reconfigurable unrolling factors to make it work well with different layers. However, such a design requires the construction of complex hardware structures, which leaves the CLP with insufficient resources for computation. Instead, we partition the available hardware resources among multiple small, specialized CLPs. Specifically, the $\textnormal{Multi-CLP}$ design adopts $G$ CLPs to process CNN layers, where $G \in [1, L]$ and $L$ is the number of CONV layers. Each CONV layer is bound to a single CLP. In other words, a fixed set of layers, referred to as $\mathfrak{L}_g$, is assigned to ${g\textnormal{-th}}$ CLP, $\mathfrak{L}_g \subseteq [L]$. Later, each CLP sequentially processes its assigned CONV layers.

In this context, we define an \textit{episode} as a single pass through CLP layers. To avoid intra-\textit{episode} data dependencies, the output FMs produced from CONV layer $\ell$ during the ${i\textnormal{-th}}$ \textit{episode} are not used in \textit{episode} $i$. Instead, they are used as input for CONV layer ${(\ell + 1)}$ in \textit{episode} ${(i + 1)}$. Therefore, CLPs need to synchronize before starting a new \textit{episode}. The advantage of $\textnormal{Multi-CLP}$ accelerator comes from various dimensions supported that can accommodate CONV layers of varying dimensions. Furthermore, by applying data pipelines to the CLP level as well, it becomes possible to work concurrently on $L$ independent images. Since a single CONV layer is processed for a given input image in an \textit{episode}, processing an image requires $L$ \textit{episodes}. Note that this nature of back-to-back processing allows data transfer for one layer to be overlapped with computation for another.

The problem now is, how to determine the appropriate number of CLPs to use, on what basis to assign a CONV layer to a CLP, and how many hardware resources each CLP can utilize? Here, we must emphasize that this research aims to optimize CNN throughput. To achieve that, each CONV layer must be assigned to the CLP that most closely matches its dimensions. Therefore, we need to look for a CONV-CLP assignment that maximizes resource utilization. Additionally, CNN throughput is constrained by the CLP that takes the longest time to finish the \textit{episode}. Thus, the \textit{episode} time for that CLP should be as short as possible. Furthermore, the size of on-chip buffers is inversely related to the required off-chip bandwidth. Therefore, we need to ensure that the $\textnormal{Multi-CLP}$ design uses sufficient on-chip buffers to minimize duplicate data transfers.

To fulfill the aforementioned requirements, in Section~\ref{sec:design_cost_and_performance}, we present an analytical and empirical model to help estimate the performance and cost, in terms of hardware resources, for a candidate $\textnormal{Multi-CLP}$ design. Then, in Section~\ref{sec:design_space_exploration}, we introduce an optimization framework that employs the proposed model as well as metaheuristics to explore the intractable design space for the optimal $\textnormal{Multi-CLP}$ design, constrained by the given CNN architecture and target FPGA specifications.

\subsection{Design Cost and Performance} \label{sec:design_cost_and_performance}

The $\textnormal{Multi-CLP}$ design for a CNN is characterized by (i)~the number of CLPs, (ii)~the assignment of CONV layers to CLPs, (iii)~the unrolling factors $\langle T_{n}, T_{m} \rangle$ of each CLP, and (iv)~the tiling parameters $\langle T_{r}, T_{c} \rangle$ of each CONV layer. Considering that synthesis, placement, and routing of a candidate $\textnormal{Multi-CLP}$ design may take several minutes or even hours, one can note that it is infeasible to perform these processes at each design point for selecting the candidate with the highest performance. Hence, this section presents an analytical and empirical design scheme for accurate modeling of cost, in terms of hardware resources used, and performance for a $\textnormal{Multi-CLP}$ design given its parameters. Here, the focus is on the key specifications and hardware resources of the target FPGA platform, including DSPs, block random access memories (BRAMs), and off-chip memory bandwidth (BW).

\subsubsection{Computational Performance} \label{sec:comp_perf}

To evaluate the efficiency of a $\textnormal{Multi-CLP}$ design, the speed of each CLP, in terms of computation cycles, must be considered. Based on Algorithm~\ref{alg:convolution_unrol_tile}, the number of cycles needed to process CONV layer $\ell$ assigned to ${g\textnormal{-th}}$ CLP is modeled as follows. Along the rows and columns dimensions of the output FMs, the algorithm iterates over the tiles and the elements in each tile. Thus, it runs for ${R_\ell \times C_\ell}$ cycles to process loop iterators $r$, $c$, $rt$, and $ct$. On the other hand, input FMs and output FMs are unrolled using the factors $T_n^{(g)}$ and $T_m^{(g)}$, respectively. Therefore, processing loop iterators $n$, $m$, $nt$, and $mt$ need ${\ceil{{N_\ell} \, / \, {T_n^{(g)}}} \times \ceil{{M_\ell} \, / \, {T_m^{(g)}}}}$ cycles. Note that the two innermost loops, highlighted in yellow, only take $1$ cycle because they are unrolled. Finally, loop iterators $i$ and $j$ are neither tiled nor unrolled, and thus, require ${K_\ell \times K_\ell}$ cycles. Altogether, the number of cycles required to process CONV layer $\ell$ on ${g\textnormal{-th}}$ CLP is calculated as 
\begin{equation}
\label{eq:CLP_CONV_cycles}
\mathrm{Comp\_Cycle}_\ell^{(g)} = \ceil*{\frac{N_\ell}{T_n^{(g)}}} \times \ceil*{\frac{M_\ell}{T_m^{(g)}}} \times R_\ell \times C_\ell \times K_\ell^2
\end{equation}

Accordingly, the computational performance of the ${g\textnormal{-th}}$ CLP design in processing CONV layer $\ell$ is defined as the total number of operations, i.e., multiply and accumulate operations, required by the CLP to process the layer over the total number of cycles needed to do so as follows
\begin{equation} \label{eq:computational_roof}
\begin{aligned}
\mathrm{Comp\_Perf}_\ell^{(g)} & = \frac{2 \times N_\ell \times M_\ell \times R_\ell \times C_\ell \times K_\ell^2}{\mathrm{Comp\_Cycle}_\ell^{(g)}}\\
&  = \frac{2 \times N_\ell \times M_\ell}{\ceil*{N_\ell \, / \, T_n^{(g)}} \times \ceil*{M_\ell \, / \, T_m^{(g)}}}
\end{aligned}
\end{equation}

\subsubsection{DSP Slice Usage}

The use of DSP slices in each CLP is dominated by the $T_m$ MAC tree tiles that work in parallel to improve computational throughput. Each MAC tree tile consists of $T_n$ parallel multipliers and an adder tree. Specifically, each multiplier performs multiplication on an input activation feature and an input kernel weight. Then, the resultant products are accumulated with the old partial result using a binary adder tree consisting of $T_n$ adders. Thus, CLP computational engine contains $T_n \times T_m$ multipliers and adders.

It is noteworthy that the number of CLP slices required depends on the operation type and data representation. To estimate how many DSP slices are needed for a given CLP design, we empirically measure the number of DSPs used to implement the multiplier and adder for each representation in the design space and use a lookup table model to calculate the total usage. Based upon this, the number of DSP slices required to implement the computational engine of ${g\textnormal{-th}}$ CLP is determined by
\begin{equation} \label{eq:DSP_usage_per_CLP}
\mathrm{DSP\_Usage}^{(g)} = \mathrm{DSP}\!\left(\mathcal{Q}_{I\!F}, \mathcal{Q}_W\right) \times T_n^{(g)} \times T_m^{(g)}
\end{equation}
where $\mathcal{Q}_{I\!F}$ and $\mathcal{Q}_W$ indicate whether input features and weights, respectively, are in single-precision floating-point representation (FP32) or quantized to low-precision representation such as $16$-bit fixed-point format (FxP16), and the function $\mathrm{DSP}(\mathcal{Q}_{I\!F}, \mathcal{Q}_W)$ returns the number of DSP slices required to implement a single multiplier and a single adder using the specified format. For instance, the FP32 adder and multiplier comprise $2$ and $3$ DSP slices, respectively~\cite{xilinx2020fp}. That is, $\mathrm{DSP(FP32, FP32)}$ equals $5$. On the other hand, the FxP16 adder and multiplier can be implemented on the same DSP slice, and therefore $\mathrm{DSP(FxP16, FxP16)}$ equals $1$. For a $\textnormal{Multi-CLP}$ design consisting of $G$ CLPs, the number of DSP slices required is the sum of the DSPs used by all CLPs
\begin{equation} \label{eq:DSP_usage_overall}
\mathrm{DSP\_Usage} = \mathlarger{\sum}_{g \, = \, 0}^{G \, - \, 1} \; \mathrm{DSP\_Usage}^{(g)}
\end{equation}

\subsubsection{BRAM Usage} \label{sec:bram_optim}

Each CLP in $\textnormal{Multi-CLP}$ design requires three on-chip buffers, namely, IF\_BUF, W\_BUF, and OF\_BUF as discussed in Section~\ref{sec:optimizing_CLP_computation_and_memory_access}. These buffers are typically constructed using BRAMs. To accurately model BRAM usage in each CLP, the following must be considered; (i)~BRAM capacity and configuration capability, (ii)~data precision, (iii)~buffer scheme (single/double buffering), (iv)~number of banks needed per buffer, (v)~number of read/write ports required for each bank, (vi)~CLP unrolling factors ($T_n$, $T_m$), and (vii)~tiling parameters ($T_r$, $T_c$) and configurations ($K$, $S$) for each CONV layer assigned to the CLP.

In this work, we employ the double-buffering scheme for all buffers to overlap off-chip data transfer with computation. Furthermore, we adopt the memory banking method to allow each buffer to provide the required number of input/output channels. Accordingly, the bank size must be large enough to support the two most demanding successive tiles. If considering the ${g\textnormal{-th}}$ CLP that processes CONV layers in subset $\mathfrak{L}_g$, the above requirement on minimum bank depth is translated as
\begin{equation} \label{eq:bank_depth}
\begin{aligned}
\mathrm{Min\_Depth}^{(g)}_d = {} \max\!\Big(& \mathrm{MFP}(d, \ell) + \mathrm{MFP}(d, \ell + 1) , \\
& \, 2 \times \mathrm{MFP}(d, \ell) \: : \; \forall \, \ell \in \mathfrak{L}_g\Big)
\end{aligned}
\end{equation}
where the function $\mathrm{MFP}(d, \ell)$ returns the memory footprint, in terms of data size, for layer $\ell$ tiles on each bank in the buffer type $d$, ${d \in \{I\!F, W, O\!F\}}$, and $\mathfrak{L}_g$ is a set of all CONV layers assigned to ${g\textnormal{-th}}$ CLP, $\mathfrak{L}_g \subseteq [L]$. The number of BRAMs per bank (BPB) is calculated by
\begin{equation} \label{eq:BRAMS_per_bank}
\mathrm{BPB}^{(g)}_d = \ceil*{\frac{\mathrm{Min\_Depth}^{(g)}_d}{\mathrm{ADDR}(\mathcal{Q}_d)}}
\end{equation}

Here, the $\mathrm{ADDR}(\mathcal{Q}_d)$ is a function that returns the appropriate BRAM depth, i.e., the number of addresses, to store data represented with the width specified by $\mathcal{Q}_d$. The BRAM model considered in this paper is based on the RAMB18E1 primitive provided by Xilinx $7$ series FPGAs~\cite{xilinx2019memory}. This memory can be configured ($depth \times width$) as a ${512 \times 36}$, ${1024 \times 18}$, ${2048 \times 9}$, ${4096 \times 4}$, ${8192 \times 2}$, or ${16384 \times 1}$. Thus, $\mathrm{ADDR(FP32)}$ is equal to $512$, while $\mathrm{ADDR(FxP16)}$ is equal to $1024$. Thus, the total number of BRAMs required to construct the buffer type $d$ of ${g\textnormal{-th}}$ CLP is calculated by
\begin{equation}
\label{eq:BRAMs_per_BUF}
\mathrm{BRAM\_Usage}_{d}^{(g)} = \mathrm{BPB}_{d}^{(g)} \times \mathcal{B}_{d}^{(g)}
\end{equation}
where $\mathcal{B}_{d}^{(g)}$ represents the number of memory banks needed to design the buffer type $d$ of ${g\textnormal{-th}}$ CLP. Details of the specification of BRAMs and the number of banks needed for each buffer type are presented next.
\\
$\mathbf{IF\_BUF}$: The CLP computational engine requires $T_n$ channels from IF\_BUF. Therefore, this buffer is designed using $T_n$ banks, i.e., $\mathcal{B}_{I\!F}^{(g)} = T_n^{(g)}$. When processing CONV layer $\ell$, its footprint on each of these banks is
\begin{equation} \label{eq:CONV_MFP_IF_BUFF}
\begin{aligned}
\mathrm{MFP}(I\!F, \ell) = {} & \big(K_\ell + S_\ell \times \left({T_r}_\ell - 1\right)\big) \times \\
& \big(K_\ell + S_\ell \times \left({T_c}_\ell - 1\right)\big)
\end{aligned}
\end{equation}

Note that each of IF\_BUF BRAMs must provide two independent ports; a read-only port for supplying CLP computational engine and a write-only port for loading off-chip memory data. Therefore, the type of these BRAMs is set to simple dual-port mode~\cite{xilinx2019memory}.
\\
$\mathbf{W\_BUF}$: The weights buffer is similar to input FMs buffer in terms of operating mode. But, unlike IF\_BUF which is shared among all MAC tree tiles, the digital multipliers of MAC tree tiles use different kernel weights, and thus require independent memory banks. Accordingly, the W\_BUF is organized into $T_n \times T_m$ banks, i.e., $\mathcal{B}_{W}^{(g)} = T_n^{(g)} \times T_m^{(g)}$. The memory footprint of CONV layer $\ell$ on each of these banks is
\begin{equation} \label{eq:CONV_MFP_W_BUFF}
\mathrm{MFP}(W, \ell) = K_\ell \times K_\ell
\end{equation}
$\mathbf{OF\_BUF}$: In each computation cycle, the old output feature is loaded and accumulated with the result of MAC tree tile. After that, the new output feature is stored back in the output FMs buffer. As we note, the accumulator needs a read port and a write port. However, these ports use the same address for both operations. Additionally, the resulting output features from the previous stage must be transferred to off-chip memory. Therefore, the OF\_BUF uses true dual-port BRAMs in write-first mode~\cite{xilinx2019memory}. One can also note that OF\_BUF must be organized into $T_m$ banks. That is, $\mathcal{B}_{O\!F}^{(g)}$ equals $T_m^{(g)}$. Thus, the memory footprint of CONV layer $\ell$ on each of these banks is calculated as
\begin{equation} \label{eq:CONV_MFP_OF_BUFF}
\mathrm{MFP}({O\!F}, \ell) = {T_r}_\ell \times {T_c}_\ell
\end{equation}

Together, the total number of BRAM resources required to construct the on-chip buffers of the $\textnormal{Multi-CLP}$ design is calculated as
\begin{equation}
\label{eq:BRAM_usage_overall}
\mathrm{BRAM\_Usage} = \mathlarger{\sum}_{g \, = \, 0}^{G \, - \, 1}\!\!\!\!\!\!\!\!\! \mathlarger{\sum}_{\;\;\;\;\;\;\;\;d \, \in \, \{{I\!F}, {W}, {O\!F}\}} \!\!\!\!\!\!\!\!\!\!\! \mathrm{BRAM\_Usage}_d^{(g)}
\end{equation}

\subsubsection{Bandwidth Usage} \label{sec:bw_optim}

The computational performance discussed in Section~\ref{sec:comp_perf} represents the maximum number of operations that can be performed with the available computational resources per cycle. Thus, it is also known as the computational roof. However, one critical problem is that CLP computation may be blocked by data transfer for memory-bound layers. Consequently, such layers cannot achieve the best performance supported by computational resources. To find the attainable computational performance in such a case, we adopt the roofline model~\cite{williams2009roofline}.

\begin{figure}[t!]
\centering
\includegraphics[width=0.49\textwidth]{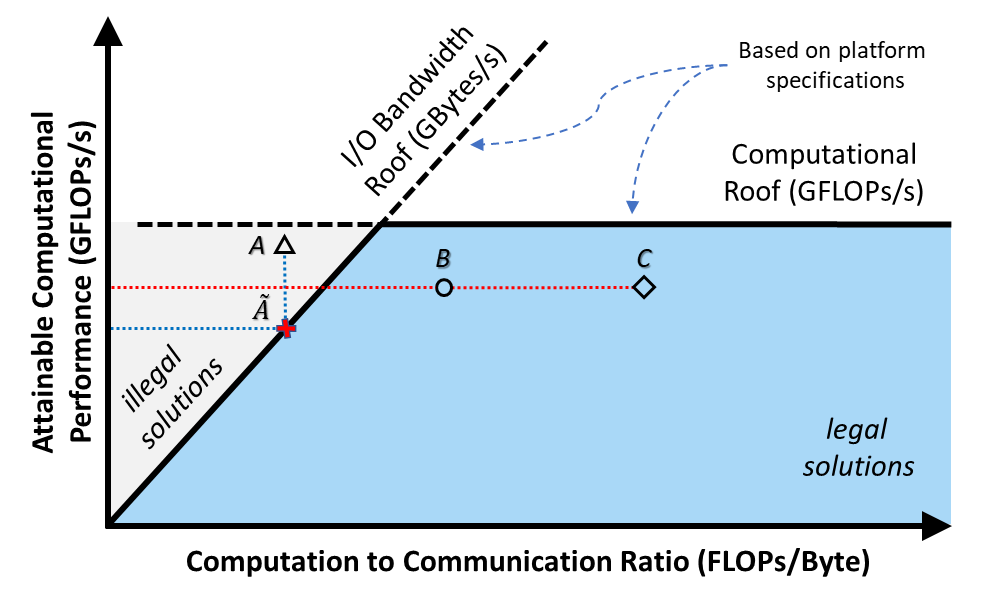}
\caption{An illustration of roofline model basis. As we can see, implementation $A$ is memory-bounded, i.e., system throughput is constrained by the communication with off-chip memory, while implementations $B$ and $C$ are computation-bounded, and thus system throughput can only be improved by using additional computational resources. One can also note that design $C$ has better data reuse compared to design $B$, and thus it is more preferable than design $B$.}
\label{fig:roofline}
\end{figure}

The roofline model is used to provide performance estimates for a candidate implementation on a particular system. It relates the attainable performance to the peak performance the system can achieve and the off-chip DRAM memory traffic as illustrated in Figure~\ref{fig:roofline}. The input/output bandwidth roof determines the maximum computational performance supported by the memory system for a given computation to communication (CTC) ratio, which is calculated as $CTC \times BW$. Thus, the attainable performance for CONV layer $\ell$ when it is processed by ${g\textnormal{-th}}$ CLP is given by
\begin{equation} \label{eq:attainable_performance}
\begin{aligned}
\mathrm{Perf}_\ell^{(g)} = \min\!\Big(\mathrm{Comp\_Perf}_\ell^{(g)},\; {CTC}_\ell^{(g)} \times BW \Big)
\end{aligned}
\end{equation}
where $\mathrm{Comp\_Perf}_\ell^{(g)}$ is the computational roof indicated in Equation~\eqref{eq:computational_roof}, 
the $BW$ refers to the off-chip memory bandwidth, and the ${CTC}_\ell^{(g)}$ describes the off-chip memory traffic that the CLP needs during CONV layer processing. The ${CTC}$ can be calculated by dividing the total number of operations by the total amount of off-chip data accessed as follows
\begin{equation}
\label{eq:CTC}
{CTC}_\ell^{(g)} = \frac{2 \times N_\ell \times M_\ell \times R_\ell \times C_\ell \times K_\ell^2}{\!\!\!\!\!\!\!\!\!\mathlarger{\sum}_{\;\;\;\;\;\;d \, \in \, \{{I\!F}, {W}, {O\!F}\}} \!\!\!\!\!\!\!\!\!\! \alpha_{d, \, \ell}^{(g)} \times \mathcal{B}_{d}^{(g)} \times \mathrm{MFP}(d, \ell)}
\end{equation}

Here, $\mathcal{B}_{d}^{(g)}$ denotes the number of memory banks in the buffer type $d$ of ${g\textnormal{-th}}$ CLP, $\mathrm{MFP}(d, \ell)$ refers to the memory footprint for CONV layer $\ell$ tiles on these memory banks as mentioned in Equations~\eqref{eq:CONV_MFP_IF_BUFF}, \eqref{eq:CONV_MFP_W_BUFF}, and \eqref{eq:CONV_MFP_OF_BUFF}, and  $\alpha_{d, \, \ell}^{(g)}$ indicates the number of off-chip memory accesses needed for buffer type $d$ during the computation of CONV layer $\ell$ by ${g\textnormal{-th}}$ CLP, such that
\begin{equation}
\label{eq:alpha}
\begin{aligned}
& \alpha_{I\!F, \, \ell}^{(g)} = \alpha_{W, \, \ell}^{(g)} = \frac{N_\ell}{T_n^{(g)}} \times \frac{M_\ell}{T_m^{(g)}} \times \frac{R_\ell}{{T_r}_\ell} \times \frac{C_\ell}{{T_c}_\ell} \\
& \alpha_{O\!F, \, \ell}^{(g)} = \frac{M_\ell}{T_m^{(g)}} \times \frac{R_\ell}{{T_r}_\ell} \times \frac{C_\ell}{{T_c}_\ell}
\end{aligned}
\end{equation}

To estimate the bandwidth required to support the maximum computation speed, we focus on the peak bandwidth that a CLP needs. Specifically, the minimum bandwidth required to achieve the peak computational performance for CONV layer $\ell$ which is processed by ${g\textnormal{-th}}$ CLP is determined by
\begin{equation}
\label{eq:min_bandwidth_for_CONV_in_CLP}
\mathrm{Min\_BW}_\ell^{(g)} = \frac{\mathrm{Comp\_Perf}_\ell^{(g)}}{{CTC}_\ell^{(g)}}
\end{equation}

For a memory bound layer, the speed of ${g\textnormal{-th}}$ CLP design in processing CONV layer $\ell$ is defined by the data
transfer cycles instead of the computation cycles mentioned in Equation~\eqref{eq:CLP_CONV_cycles}. Thus, the arithmetic utilization of $\textnormal{Multi-CLP}$ design processes a CNN with $L$ CONV layers using $G$ CLPs is given by
\begin{equation} \label{eq:utilization}
\mathrm{Utilization} = \frac{\mathlarger{\sum}_{g \, \in \, [G]} \;\; \mathlarger{\sum}_{\ell \, \in \, \mathfrak{L}_g} \; \lambda_\ell^{(g)} \times \mathrm{Cycle}_\ell^{(g)}}{G \times \max\!\left(\mathrm{Cycle}^{(g)} \: : \; \forall \, g \in [G]\right)}
\end{equation}
where $\lambda_\ell^{(g)}$ is the utilization ratio of ${g\textnormal{-th}}$ CLP computational resources while it is processing CONV layer $\ell$ that is computed as mentioned in Equation~\eqref{eq:CONV_utilization}.

\section{Design Space Exploration} \label{sec:design_space_exploration}

In this section, we discuss the difficulty of determining the value of design variables for multiple CLPs used to accelerate CNNs on FPGAs. Then, we present our proposed approach to finding the optimal $\textnormal{Multi-CLP}$ design for a given CNN architecture and resource budget.

\subsection{Problem Statement and Formulation}

The implementation of a CNN architecture on an FPGA platform using multiple CLPs while optimizing their performance is a non-trivial task. This is because it requires optimizing the utilization of both processing units and memory units to minimize the execution cycles. However, the number of cycles required depends on the number of CLPs used, the assignment of CONV layers to CLPs, the loop unrolling factors $\langle T_{n}, T_{m} \rangle$ for each CLP, and the tilling parameters $\langle T_{r}, T_{c} \rangle$ for each CONV layer, which imposes an intractable design space. Furthermore, we must ensure that the peak memory bandwidth and resources utilized for a given candidate design comply with the given hardware constraints. 

Thus, determining the best values of the design variables through exhaustive search is infeasible, especially when the number of variables is large and/or FPGA resources are limited. Additionally, the number of possible assignments for CONV layers to CLPs is exponential in the number of layers. Considering the unrolling factors $\langle T_{n}, T_{m} \rangle$ as a motivating example, the search space of a given CNN structure with $L$ CONV layers would consist of ${L^L \times \max\!\left(N_\ell : \forall \, \ell \in [L]\right)^L \times \max\!\left(M_\ell : \forall \, \ell \in [L]\right)^L}$ possible solutions, where $N_\ell$ and $M_\ell$ are the number of input FMs and output FMs for CONV layer $\ell$, respectively.

For such a computationally hard problem, there is no existing deterministic algorithm that can find the optimal solution.
Ad-hoc heuristic solutions have been employed by Shen et al.~\cite{shen2017maximizing} to find good solutions. However, iterative non-deterministic heuristics can be employed to efficiently traverse the search space as they have been proven to find excellent solutions to such hard problems~\cite{sait1999iterative}. In this work, we employ simulated annealing (SA) and tabu search (TS) algorithms to find the optimal $\textnormal{Multi-CLP}$ design for a given CNN structure. The SA and TS are adopted due to their efficiency in finding high-quality solutions to such kinds of problems, simplicity of implementation, ability to provide a globally optimal solution in a reasonable time, the ease of tuning their parameters, and providing a balance between the exploration of solution space and the exploration of information obtained~\cite{sait1999vlsi}.

The space of solutions comprises finding optimal values for the following four elements; (i)~the number of CLPs to be used, (ii)~the assignment of CONV layers to CLPs, (iii)~the values of $T_n$ and $T_m$ for each CLP, (iv)~the values of $T_r$ and $T_c$ for each CONV layer. The solution representation is composed of $L+2$ rows as shown in Table~\ref{table:assignment-example}. Each one of the first $L$ rows represents a single CLP, while the last two rows represent the $T_r$ and $T_c$ values of each CONV layer. On the other hand, CLP rows contain $L+2$ columns. The first two columns store the current values of $T_n$ and $T_m$, while the remaining $L$ columns indicate the assignment of layers to CLPs by storing $0$ (unassigned) or $1$ (assigned). Here, we must emphasize that a column representing a layer assignment must have a single $1$, while a row can have multiple $1$s.

Initially, the search begins with a random assignment of CONV layers to CLPs, and random $T_n$ and $T_m$ values for each CLP provided they do  not violate the hardware resource constraints. The candidate solution represented in Table~\ref{table:assignment-example} shows that the total number of DSPs is $100$, which can be calculated based on Equation~\eqref{eq:DSP_usage_overall} assuming features and weights are in $32$-bit floating-point representation. One can notice that $CLP_4$ and $CLP_5$ do not require any DSP since no layer is assigned to them. Note that in the case where  more than one layer is assigned to a CLP, the CLP processes these layers sequentially one after another. Thus, the number of cycles required by a CLP is the sum of the number of cycles needed to process all its layers. 

\begin{table}[!t]
\centering
\caption{An example of assigning 5 CONV layers to 5 CLPs with initial random values to $T_n$, $T_m$, $T_r$, and $T_c$.}
\resizebox{\columnwidth}{!}{%
\begin{tabular}{|c||c|c||c|c|c|c|c||c|} 
\hline
 \textbf{Config} & $\bm{T_n}$ & $\bm{T_m}$ & $\bm{L_1}$ & $\bm{L_2}$ & $\bm{L_3}$ & $\bm{L_4}$ & $\bm{L_5}$ & \textbf{\# DSPs} \\ 
\hline
\hline
 $\bm{CLP_1}$ & 3 & 5 & 0 & 0 & 0 & 1 & 1 & 75 \\ 
\hline
 $\bm{CLP_2}$ & 4 & 1 & 0 & 1 & 0 & 0 & 0 & 20 \\ 
\hline
 $\bm{CLP_3}$ & 1 & 1 & 1 & 0 & 1 & 0 & 0 & 5 \\ 
\hline
 $\bm{CLP_4}$ & 7 & 2 & 0 & 0 & 0 & 0 & 0 & - \\ 
\hline
 $\bm{CLP_5}$ & 3 & 6 & 0 & 0 & 0 & 0 & 0 & - \\ 
\hline
\hline
\multicolumn{3}{|c||}{$\bm{T_r}$} & \multicolumn{1}{c|}{3} & \multicolumn{1}{c|}{5} & \multicolumn{1}{c|}{2} & \multicolumn{1}{c|}{4} & \multicolumn{1}{c||}{1} & \multicolumn{1}{c|}{}  \\ 
\cline{1-8}
\multicolumn{3}{|c||}{$\bm{T_c}$} & \multicolumn{1}{c|}{3} & \multicolumn{1}{c|}{3} & \multicolumn{1}{c|}{3} & \multicolumn{1}{c|}{4} & \multicolumn{1}{c||}{2} & \multicolumn{1}{c|}{} \\
\hline

\end{tabular}%
}
\label{table:assignment-example}
\end{table}

The proposed optimization algorithm explores the design space by generating neighboring solutions from the current one, aiming to reach the optimal candidate. In our formulation, a neighboring solution is generated by performing a single move from the current solution. A move or step is defined as a single change in the current solution either by changing the layer assignment to a different CLP, or by changing one of the $T_n$, $T_m$, $T_r$, or $T_c$ parameters for a CLP. An optimal solution is defined as a $\textnormal{Multi-CLP}$ design that requires the least number of cycles to process all layers. The number of cycles required for the candidate design $\mathcal{S}$ is calculated as
\begin{equation} \label{eq:design_cycles}
\mathrm{Cycle}\left(\mathcal{S}\right) = \max\!\left(\mathlarger{\sum}_{\ell \, \in \, \mathfrak{L}_g} \; \mathrm{Cycle}_\ell^{(g)} \: : \; \forall \, g \in [G]\right)
\end{equation}
where $\mathrm{Cycle}_\ell^{(g)}$ is the number of cycles required to process CONV layer $\ell$ in ${g\textnormal{-th}}$ CLP, $G$ is the number of CLPs in the candidate solution $\mathcal{S}$, and $\mathfrak{L}_g$ is a set of all CONV layers assigned to ${g\textnormal{-th}}$ CLP, $\mathfrak{L}_g \subseteq [L]$. Note that the number of cycles for memory-bounded layers is defined by the data transfer cycles, while the number of cycles for computational-bounded layers is defined by the computational roof as discussed in Section~\ref{sec:design_cost_and_performance}. Based on the discussion above, the optimization problem can be summarized as 
\begin{equation} \label{eq:optimization}
\begin{aligned}
\;\;\;\;\;\; & \underset{\mathcal{S}}{\text{minimize}}
& & \!\!\!\!\!\!\!\!\!\!\! \mathrm{Cycle}\left(\mathcal{S}\right)\\
& \;\;\;\;\;\; \text{subject to}
&& \mathrm{DSP\_Usage}_\mathcal{S} \leq \mathrm{DSP}_{max}\\
& && \mathrm{BRAM\_Usage}_\mathcal{S} \leq \mathrm{BRAM}_{max}\\
\end{aligned}
\end{equation}

Here, the $\mathrm{DSP\_Usage}_\mathcal{S}$ and $\mathrm{BRAM\_Usage}_\mathcal{S}$ are the total number of DSP and BRAM resources required for the solution $\mathcal{S}$, which are computed as discussed in Equations~\eqref{eq:DSP_usage_overall}~and~\eqref{eq:BRAM_usage_overall}, respectively, the $\mathrm{DSP}_{max}$ and $\mathrm{BRAM}_{max}$ are the number of DSPs and BRAMs available in the target FPGA platform. For each candidate solution, the memory optimization to find the values of $T_r$ and $T_c$ for each CONV layer is based on minimizing the peak memory bandwidth required as discussed in Section~\ref{sec:bw_optim}. The parameters $T_r$ and $T_c$ in turn would set the buffer sizes for each CLP, and thus they must comply with the available on-chip memory budget as demonstrated in Section~\ref{sec:bram_optim}.

\subsection{Optimization of Multi-CLP Designs}

In the section, we describe the employment of simulated annealing and tabu search metaheuristic algorithms in optimizing the implementation of FPGA-based CNN accelerators based on multiple CLPs.

\subsubsection{Simulated Annealing (SA)}

The SA is an effective optimization tool  based on inspiration derived from       annealing of metals~\cite{kirkpatrick1983optimization, vcerny1985thermodynamical}. Its  power lies in its simplicity of implementation, and the ease of tuning its parameters. During search, the moves that cause a decrease in the cost function, also known as good moves, are always accepted. On the other hand, the moves that lead to higher cost, also known as bad moves or uphill moves, are sometimes accepted with a probability that depends on a parameter called temperature $(T)$. Initially, that is when $T$ is high, the search is almost random. As $T$ decreases, the probability of accepting bad moves decreases and the search begins to become greedy. At zero temperature, the search becomes totally greedy, and only good moves are accepted.

The basic structure of SA algorithm is shown in Algorithm~\ref{alg:sa},  the core of which  is the $\Metropolis$ procedure. The $\Metropolis$ procedure  simulates the annealing process at a given temperature value $T$~\cite{metropolis1953equation}. It receives the current temperature $T$ and the current solution $\mathcal{S}$ as input and improves it. It is also provided with the value $M$, which is a multiplication factor that increases the amount of time annealing is applied at a given temperature $T$~\cite{youssef2001evolutionary}. The temperature is initialized to a value $T_0$ (explained below), and is slowly decreased  using  the parameter $\alpha$. As temperature is lowered, the amount of time spent in annealing at a given temperature is gradually increased, using the parameter $\beta$, $\beta > 1$. The  time spent in each call to $\Metropolis$ procedure is stored in variable $Time$. SA algorithm stops when the time limit is reached, i.e., $Time \geq MaxTime$.

\begin{algorithm}[!t]
\small
\SetAlgoLined
\DontPrintSemicolon
\KwIn{${\textnormal{The initial solution }(\mathcal{S}_{0})\textnormal{, initial temperature }(T_{0})\textnormal{,}}$ ${\textnormal{cooling rate }(\alpha)\textnormal{, annealing temperature constant}}$ ${(\beta)\textnormal{, total allowed time }(MaxTime)\textnormal{, and time until}}$ ${\textnormal{next  update }(M)}$.} 
\KwOut{The best admissible solution ($\mathcal{S}^{*}$).}
\SetKwFunction{SA}{SA}
\SetKwProg{Fn}{Procedure}{:}{}
\nonl\SetAlgoNoLine\Fn{\SA{$\mathcal{S}_{0}, T_{0}, \alpha, \beta, MaxTime, M$}}{
\SetAlgoLined
\tikzmk{A}{$\mathcal{S} \gets \mathcal{S}_{0}$

    $T \gets T_{0}$
        
    $Time \gets 0$
    
    \Repeat{$Time \geq MaxTime$}{
        $\mathcal{S} \gets \Metropolis\!\left(\mathcal{S}, T, M\right)$
        
         $Time \gets Time + M$
         
         $T \gets \alpha \times T$
         
         $M \gets \beta \times M$
    }
    $\mathcal{S}^{*} \gets \mathcal{S}$

}\tikzmk{B}\boxitLL{red}
\KwRet $\;\mathcal{S}^{*}$
}

\SetKwProg{Fn}{Procedure}{:}{}
\nonl\SetAlgoNoLine\Fn{\Metropolis{$\mathcal{S}, T, M$}}{
\SetAlgoLined
\tikzmk{A}{\Repeat{$M = 0$}{
        $\mathcal{S}_{new} \gets \Neighbor\!\left(\mathcal{S}\right)$ \color{DarkBlue}{\Comment{\text{\scriptsize Do a single move}}}
        
         $\Delta h \gets \Cost\!\left(\mathcal{S}_{new}\right) - \Cost\!\left(\mathcal{S}\right)$ \color{DarkBlue}{\Comment{\text{\scriptsize Using Eq.~\eqref{eq:design_cycles}}}}
         
         \If{$(\Delta h < 0)$ {\bf or} $(random < e^{-\Delta h / T})$}{$\mathcal{S} \gets \mathcal{S}_{new}$ \color{DarkBlue}{\Comment{\text{\scriptsize Accept the solution}}}
         }
         
         $M \gets M - 1$
    }
}\tikzmk{B}\boxitLL{blue}
\KwRet $\;\mathcal{S}$

}
\caption{Simulated Annealing Algorithm.}
\label{alg:sa}
\end{algorithm}

The $\Metropolis$ procedure generates a local new solution $\mathcal{S}_{new}$ from a given solution $\mathcal{S}$
using the procedure $\Neighbor$. Two perturbation schemes are used to generate a new neighbor out of the current solution; either (i)~by randomly changing the assignment of a single CONV layer to CLPs, or (ii)~by mutating $T_n$ or $T_m$ of a randomly selected CLP. These two perturbations are performed probabilistically,  by making changes to $T_n$ or $T_m$ with $80$\% probability while making changes of CONV layer assignment to CLPs  with $20$\% probability. The rationale for this is that  because of the observation that optimizing the values of $T_n$ and $T_m$ for  CLPs has greater effect on the solution cost. Additionally, during later iterations,  changing CONV layer assignment to CLPs results in more invalid solutions. This is because adding new CLPs would require additional DSPs which may exceed the DSP constraint.

The function $\Cost\!\left(\mathcal{S}\right)$ returns the cost of a given solution $\mathcal{S}$, which represents the number of cycles required to complete the execution for the given solution representation as mentioned in Equation~\eqref{eq:design_cycles}. The new solution is accepted if its cost, $\Cost\!\left(\mathcal{S}_{new}\right)$, is better than the
cost of the current solution $\mathcal{S}$. The new solution is also accepted by $\Metropolis$ probabilistically when its cost is higher than the cost of the current solution.
This is done by generating a random number in the range $0$ to $1$. If the generated number is smaller than $(e^{-{\Delta}h/T} )$, then the inferior solution is accepted, resulting in hill-climbing, i.e., transition from a low-cost solution to a high-cost solution, where $e$ is the Euler's number, and ${\Delta}h$ is the change in cost.
 
\subsubsection{Tabu Search (TS)} 

The TS is an another non-deterministic algorithm that also starts from an  initial feasible solution and generates a list of candidate solutions, that forms what is known as the candidate list ($\mbox{V}^{*}$), by making neighborhood moves, and then selecting from this list the solution   that is  best   among all candidates   in  the current iteration~\cite{glover1989tabu, glover1990tabu, glover1993user, glover1998tabu}. To avoid move reversals, that is recycling back to previously visited solutions,  a device called tabu list (TL) is used that stores some attribute(s) of these moves. The TL has a given  size and can be viewed as a window or queue on accepted moves. The moves stored in TL are not allowed as they may undo previous moves returning back to the same solution.

The tabu status is overridden when certain criteria, known as aspiration criteria (AC), are satisfied. The AC temporarily override the tabu status if the move causes the cost to be less than the aspiration level (AL), a value defined based on the AC adopted. For instance, the “best cost” criterion defines the AL as the cost of the best admissible solution. Accordingly, if the tabu move leads to a solution whose cost is less than AL, which is an indication that there is no cycling back to a previously visited solution, then the AC override the tabu status and accept the move. Additionally, the AL is updated to the cost of the new solution.

\begin{algorithm}[!t]
\small
\SetAlgoLined
\DontPrintSemicolon
\KwIn{${\textnormal{A set of feasible solutions }(\Omega)\textnormal{, candidate list size}}$ ${(N)\textnormal{, tabu list size }(M)\textnormal{, aspiration criteria }(AC)\textnormal{,}}$ ${\textnormal{total allowed time }(MaxTime)}$.} 
\KwOut{The best admissible solution ($\mathcal{S}^{*}$).}
\SetKwFunction{TS}{TS}
\SetKwFunction{Neighbor}{Neighbor}
\SetKwProg{Fn}{Procedure}{:}{}
\nonl\SetAlgoNoLine\Fn{\TS{$\Omega, N, M, AC, MaxTime$}}{
\SetAlgoLined
\tikzmk{A}{Start with an initial feasible solution $\mathcal{S}$, $\mathcal{S} \in \Omega$

Initialize best admissible solution and current time

\nonl $\;\;\;\mathcal{S}^{*} \gets \mathcal{S}$ 

\nonl $\;\;\;Time \gets 0$

Initialize tabu list and aspiration level

\nonl $\;\;\;{\bf TL} \gets \{\}$ 

\nonl $\;\;\;{\bf AL} \gets \Cost\!\left(\mathcal{S}^{*}\right)$\color{DarkBlue}{\Comment{\text{\scriptsize Assuming {\it AC} is {\it "best cost"}}}}
    
    \Repeat{$Time = MaxTime$}{
        Generate neighborhood solutions of $\mathcal{S}$, $\aleph\!\left(\mathcal{S}\right)$

        Select $N$ neighbor solutions and add them to candidate list
        
        \nonl $\;\;\;\;\;\;\;\mbox{\bf V}^{\bf *} \gets \Sample\!\left(\aleph\!\left(\mathcal{S}\right), N\right)$
        
        Choose best solution in candidate list
        
        \nonl $\;\;\;\;\;\;\;\mathcal{S}^{*} \gets \Best\!\left(\mbox{\bf V}^{\bf *}\right)$
        
        \If{$\Move\!\left(\mathcal{S}, \mathcal{S}^{*}\right) \notin$ {\bf TL} {\bf or} $\Cost\!\left(\mathcal{S}^{*}\right) <$ {\bf AL}}{Update tabu list
    
            \nonl $\;\;\;{\bf TL} \gets {\bf TL}[1:M] + \{\Move\!\left(\mathcal{S}^{*}, \mathcal{S}\right)\}$ 
            
            $\mathcal{S} \gets \mathcal{S}^{*}$ \color{DarkBlue}{\Comment{\text{\scriptsize Accept the solution}}}
        
            \If{$\Cost\!\left(\mathcal{S}\right) <$ {\bf AL}}{$\!\!\!{\bf AL} \gets \Cost\!\left(\mathcal{S}\right)$ \color{DarkBlue}{\Comment{\text{\scriptsize Update aspiration level}}}}
        
            $Time \gets Time + 1$
         }
    }

}\tikzmk{B}\boxitLL{yellow}
\KwRet $\;\mathcal{S}^{*}$
}
\caption{Tabu Search Algorithm.}
\label{alg:ts}
\end{algorithm}

An algorithmic description of TS metaheuristic is shown in Algorithm~\ref{alg:ts}. The procedure  starts from a feasible 
solution, a neighborhood $\aleph\!\left(\mathcal{S}\right)$ is defined for each solution $\mathcal{S}$. Among all the possible neighborhood solutions, $N$ neighbor solutions are randomly selected and added to the candidate list using the procedure $\Sample\!\left(\aleph\!\left(\mathcal{S}\right), N\right)$. Note that the number of possible neighborhood solutions is typically much larger than the candidate list size, i.e., ${  | \mbox{V}^{*} |  = N \ll   | \aleph\!\left(\mathcal{S}\right) |}$. From these $N$ neighborhood  solutions in the candidate list, the best solution is chosen for consideration as the next solution  $\mathcal{S}^{*}$. The selected solution could have a cost higher than the cost of the current solution resulting in hill-climbing.

Two perturbation schemes are  used to generate neighborhood solutions; either (i)~by randomly changing the assignment of a single CONV layer to CLPs, or (ii)~by mutating $T_n$ or $T_m$ of a randomly selected CLP. For each of the perturbation schemes, a separate TL is maintained. The attributes stored in TLs are related to the perturbation scheme. In the case where a perturbation consists of changing the assignment of a CONV layer to CLPs, the CONV layer ID and the CLP ID to which it is assigned are stored. For the case of changing parameter $T_n$ or $T_m$, the attributes saved are the CLP ID and the new value of the parameter. These two perturbations are done with the same probabilities as in the case of SA, i.e., $80$\% for changing $T_n$ or $T_m$, and $20$\% for changing CONV layer assignment to CLPs.

\section{Experiments and Results} \label{sec:results}

In this section, we describe our experimental settings. Then, we employ the proposed optimization framework to design multiple CLPs, $\textnormal{Multi-CLP}$ in short, to accelerate four widely used typical CNNs on FPGAs. Next, we discuss the results and findings. Additionally, we compare the $\textnormal{Multi-CLP}$ accelerator designed using the presented metaheuristic-based framework with the state-of-the-art $\textnormal{Single-CLP}$ and $\textnormal{Multi-CLP}$ design frameworks.

\subsection{Experimental Settings}

To demonstrate the versatility of metaheuristics in designing $\textnormal{Multi-CLP}$ accelerators for CNNs, we conduct extensive experiments on the ImageNet large-scale visual recognition challenge (ILSVRC) dataset~\cite{russakovsky2015imagenet}, which is known as one of the most popular image classification benchmarks. The ILSVRC dataset contains $50$ thousand images for validation, all of which are natural high-resolution images. Each image is annotated as one of $1,000$ classes. To meet the configurations required for input data to CNNs, images are resized and then center cropped before being used as input to the network.

For those experiments, we use our metaheuristic-based optimization technique discussed in Section~\ref{sec:design_space_exploration} to design $\textnormal{Multi-CLP}$ accelerators for four benchmark architectures, namely, AlexNet~\cite{krizhevsky2012imagenet}, SqueezeNet~$1.1$~\cite{iandola2016squeezenet}, VGGNet~\cite{simonyan2014very}, and GoogLeNet~\cite{szegedy2015going}, as they are well-known in the field of image classification. The number of CONV layers in AlexNet, VGGNet, SqueezeNet~$1.1$, and GoogLeNet architectures is $10$, $13$, $26$, and $57$, respectively. To investigate the applicability of the proposed framework in designing $\textnormal{Multi-CLP}$ accelerators for CNNs with different bit-precision levels, we consider two formats; $32$-bit floating-point and $16$-bit fixed-point.

\begin{figure*}[t!]
\centering
\includegraphics[width=0.95\textwidth]{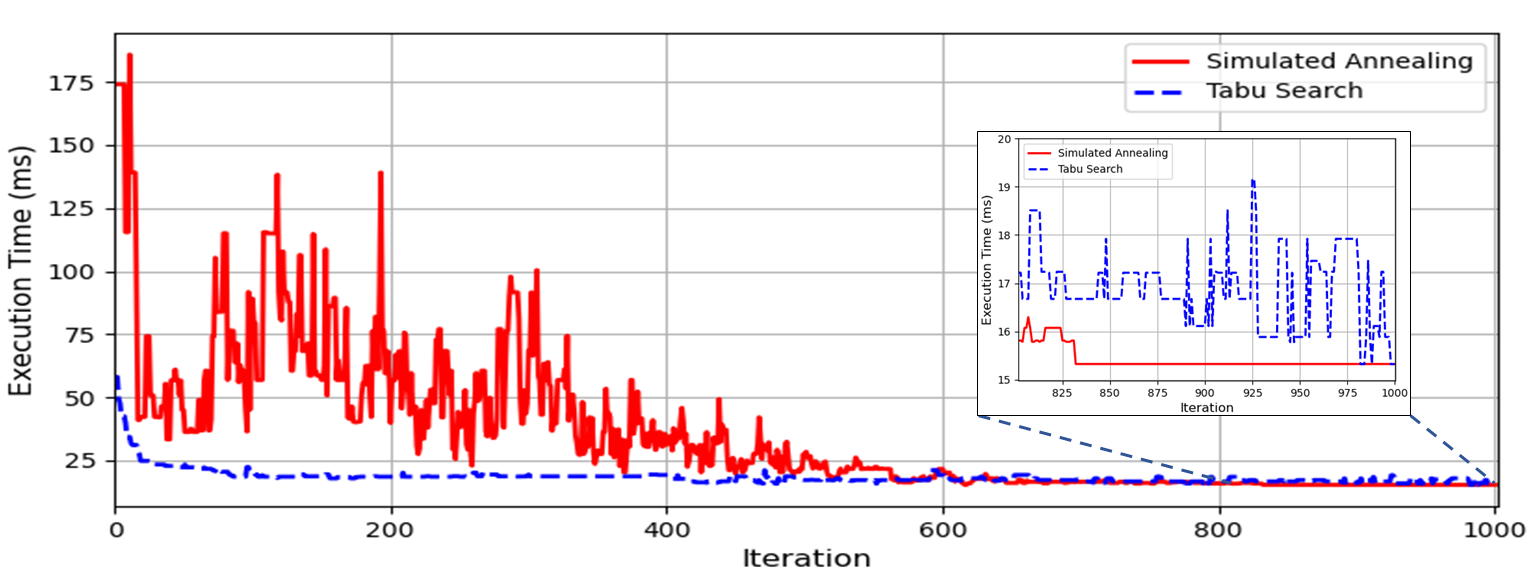}
\caption{Cost function optimization progress for a run of $1,000$ iterations for AlexNet on 485T FPGA using simulated annealing and tabu search metaheuristic algorithms.}
\label{fig:SA_TS}
\end{figure*}

The $\textnormal{Multi-CLP}$ design is implemented as a parameterized Verilog hardware description language (Verilog-HDL). The design is synthesized and implemented using Xilinx Vivado Design Suite (v2020). Our application targets the VC707 and VC709 FPGA boards, both boards operate at a frequency of $100$ MHz but each has different hardware resources. Table~\ref{tab:fpga_board_resources} summarizes the hardware specifications of these two boards in terms of FPGA platform, DSPs, $18$ Kb BRAM (BRAM18K), LUTs, FFs, and external memory.

The optimization problem described in Equation~\eqref{eq:optimization} employs the computational performance and resource utilization models discussed in Section~\ref{sec:design_cost_and_performance} to find the highest-throughput $\textnormal{Multi-CLP}$ design. The maximum number of DSPs ($\mathrm{DSP}_{max}$) and BRAMs ($\mathrm{BRAM}_{max}$) available for the optimization process is set to $80$\% of the board’s resources as in~\cite{shen2017maximizing}. This is due to the need for some hardware resources to implement, for example, the soft-core processor and controllers. Accordingly, the $\mathrm{DSP}_{max}$ and $\mathrm{BRAM}_{max}$ are set to $2,240$ and $1,648$ for the VC707 board, respectively, while maximum of $2,880$ DSPs and $2,352$ BRAMs are allowed for the optimization procedure when working on the VC709 board.

With regards to the optimization procedure, the SA metaheuristic algorithm needs to start from a high temperature $T$, which is set in our experiments  to $25,000$. The parameter $\alpha$ for updating the temperature is set to the value $0.99$. On the other hand, the constant $\beta$ which controls the time spent at each annealing temperature is set to $1.005$.  The stopping criterion is set to $1,000$ iterations. For TS metaheuristic algorithm, the size of the candidate list used is $20$, as it has been found based on experimental analysis that increasing this number does not make significant improvements while decreasing it has a negative effect on the quality of the obtained solutions. Based on the empirical analysis, the size of the tabu list chosen was $7$, as this gave the best results, while  the stopping criterion is set to $1,000$ iterations. For each performed experiment, we have reported the best results obtained from $10$ runs.

\begin{table}[t!]
\renewcommand{\arraystretch}{1.4}
\begin{center}
\captionsetup{justification=centering}
\caption{FPGA boards hardware resources.}
\label{tab:fpga_board_resources}
\footnotesize
\begin{tabular}{| l | c | c |}
\hline
\textbf{Specification} & \textbf{VC707~\cite{xilinx2019vc707}} & \textbf{VC709~\cite{xilinx2019vc709}}  \\
\hline
\hline
\multicolumn{1}{| l |}{\multirow{2}{*}{\textbf{Platform}}} & \multicolumn{1}{ c |}{\multirow{2}{*}{\makecell{Virtex-7 \\ VX485T}}} & \multicolumn{1}{ c |}{\multirow{2}{*}{\makecell{Virtex-7 \\ VX690T}}} \\
 &  &  \\
\hline
\textbf{DSP Slices} & $2,800$ & $3,600$ \\
\hline
\textbf{BRAM18K} & $2,060$ & $2,940$ \\
\hline
\textbf{LUTs} & $303,600$ & $433,200$ \\
\hline
\textbf{FFs} & $607,200$ & $866,400$ \\
\hline
\multicolumn{1}{| l |}{\multirow{2}{*}{\makecell{\textbf{External} \\ \textbf{Memory}}}} & \multicolumn{1}{ c |}{\multirow{2}{*}{\makecell{$1$GB DDR3 \\ SODIMM}}} & \multicolumn{1}{ c |}{\multirow{2}{*}{\makecell{$2 \times 4$GB DDR3 \\ SODIMM}}}\\
 &  &  \\
\hline
\end{tabular}
\end{center}
\end{table}

\subsection{Results and Discussion}

In the first experiment, we study the effectiveness of the proposed optimization framework in finding the optimal configurations that maximize the throughput of the $\textnormal{Multi-CLP}$ design. Hence, this experiment uses the computational performance and resource usage models as well as the maximum resources available on the FPGA to find the solution with the minimum execution cycles.

The behavior of a single run of SA algorithm for AlexNet CNN acceleration using multiple CLPs on 485T FPGA is shown in Figure~\ref{fig:SA_TS}. The figure shows the variation in the execution time, which is defined as the total number of cycles divided by the operating frequency. The number of cycles is the cost function, illustrated in Equation~\eqref{eq:design_cycles}, that SA algorithm tries to optimize for $1,000$ iterations when run with the parameters mentioned in the previous section. We refer to this approach as $\textnormal{SA-based}$ $\textnormal{Multi-CLP}$. As can be seen, since the algorithm starts with a high temperature, the fluctuation in costs is large, which gets reduced over time as the temperature is reduced and better solutions are discovered with minimal cost variations toward the end. Final solution configurations can be found in Table~\ref{tab:AlexNet_fp32_485T}.

Similarly, the behavior of a single run of TS algorithm for the same CNN and the same FPGA board is also illustrated. This approach to designing a $\textnormal{Multi-CLP}$ based on TS algorithm is indicated by the $\textnormal{TS-based}$ $\textnormal{Multi-CLP}$. As can be seen from Figure~\ref{fig:SA_TS}, the cost reduces very quickly in the first few iterations as there are several neighboring candidates to choose from in each iteration, and then, it reduces at a slower rate in subsequent iterations with hill-climbing behavior, until reaching the final solution shown in Table~\ref{tab:AlexNet_fp32_485T}. 

From the figure, we can deduce the effectiveness of these algorithms in finding the optimal solution to the $\textnormal{Multi-CLP}$ design problem. Both algorithms were able to efficiently explore the intractable design space and find $\textnormal{Multi-CLP}$ design solutions with approximately the same low level of execution time, but each has a different final solution. It is noteworthy that the optimization framework is very fast, it explores the search space in less than a minute on a general-purpose processor.

\begin{table*}[t!]
\renewcommand{\arraystretch}{1.4}
\begin{center}
\captionsetup{justification=centering}
\caption{Single-CLP and Multi-CLP accelerators for AlexNet on 485T FPGA using 32-bit floating-point data.}
\label{tab:AlexNet_fp32_485T}
\footnotesize
\begin{tabular}{| l || c || >{\centering}p{0.52in} | >{\centering}p{0.52in} || c || c || c || c || c |}
\hline
\multirow{2}{*}{\textbf{Technique}} & \multirow{2}{*}{\textbf{CLP No.}} & \multicolumn{2}{ c ||}{\multirow{1}{*}{\textbf{\textbf{Unrolling Factors}}}} & \multicolumn{1}{ c ||}{\multirow{2}{*}{\makecell{\textbf{Layer} \\ \textbf{Mapping}}}} & \multicolumn{1}{ c ||}{\multirow{2}{*}{\textbf{Cycles}}} & \multicolumn{1}{ c ||}{\multirow{2}{*}{\makecell{\textbf{Execution} \\ \textbf{Time (ms)}}}} & \multicolumn{1}{ c ||}{\multirow{2}{*}{\textbf{DSPs}}} & \multicolumn{1}{ c |}{\multirow{2}{*}{\textbf{BRAMs}}} \\
\cline{3-4}
& & $\mathbfit{T_n}$ & $\mathbfit{T_m}$ & & & & & \\
\hline
\hline
\multirow{5}{*}{\makecell{\textbf{Single-CLP} \\ \textbf{\cite{zhang2015optimizing}}}} & \multirow{5}{*}{$CLP_1$} & \multirow{5}{*}{$7$} & \multirow{5}{*}{$64$} & $1a \;\; , \;\; 1b$ & $\mathbf{732 \times 10^3}$ & \multirow{5}{*}{$20.06$} & \multirow{5}{*}{$2,240$} & \multirow{5}{*}{$618$} \\
\cdashline{5-6}
 & &  & & $2a \;\; , \;\; 2b$ & $\mathbf{510 \times 10^3}$ & & & \\
 \cdashline{5-6}
  & &  & & $3a \;\; , \;\; 3b$ & $\mathbf{338 \times 10^3}$ & & & \\
  \cdashline{5-6}
   & &  & & $4a\;\; , \;\; 4b$ & $\mathbf{256 \times 10^3}$ & & & \\
   \cdashline{5-6}
    & &  & & $5a \;\; , \;\; 5b$ & $\mathbf{170 \times 10^3}$ & & & \\
\hline
\multirow{5}{*}{\makecell{\textbf{Multi-CLP} \\ \textbf{\cite{shen2017maximizing}}}} & $CLP_1$ & $3$ & $24$ & $1a \;\; , \;\; 1b$ & $1,464 \times 10^3$ & \multirow{5}{*}{$15.58$} & \multirow{5}{*}{$2,240$} & \multirow{5}{*}{$731$} \\
\cdashline{2-6}
 & $CLP_2$ & $8$ & $19$ & $2a \;\; , \;\; 2b$ & $1,531 \times 10^3$ & & & \\
\cdashline{2-6}
  & $CLP_3$ &  $1$ & $96$ & $3a \;\; , \;\; 3b$ & \textit{$\mathbf{1,558 \times 10^3}$} & & & \\
 \cdashline{2-6}
   & \multirow{2}{*}{$CLP_4$} &  \multirow{2}{*}{$2$} & \multirow{2}{*}{$64$} & $4a\;\; , \;\; 4b$ & $876 \times 10^3$ & & & \\
   \cdashline{5-6}
    &  &  & & $5a \;\; , \;\; 5b$ & $584 \times 10^3$ & & & \\
\hline
\multirow{8}{*}{\makecell{\textbf{SA-based}\\ \textbf{Multi-CLP} \\ {(ours)}}} & \multirow{2}{*}{$CLP_1$} &  \multirow{2}{*}{$3$} & \multirow{2}{*}{$24$} & $1a$ & $732 \times 10^3$ & \multirow{8}{*}{$15.31$} & \multirow{8}{*}{$2,240$} & \multirow{8}{*}{$644$} \\
\cdashline{5-6}
    & &  & & $4a$ & $779 \times 10^3$ & & & \\
\cdashline{2-6}
 & \multirow{2}{*}{$CLP_2$} & \multirow{2}{*}{$3$} & \multirow{2}{*}{$24$} & $1b$ & $732 \times 10^3$ & & & \\
 \cdashline{5-6}
  & &  & & $4b$ & $779 \times 10^3$ & & & \\
 \cdashline{2-6}
   & \multirow{2}{*}{$CLP_3$} & \multirow{2}{*}{$16$} & \multirow{2}{*}{$11$} & $2a \;\; , \;\; 2b$ & $\mathbf{1,312 \times 10^3}$ & & & \\
  \cdashline{5-6}
    & &  & & $5a$ & $\mathbf{219 \times 10^3}$ & & & \\
\cdashline{2-6}
    & \multirow{2}{*}{$CLP_4$} & \multirow{2}{*}{$16$} & \multirow{2}{*}{$8$} & $3a \;\; , \;\; 3b$ & $1,168 \times 10^3$ & & & \\
\cdashline{5-6}
    & &  & & $5b$ & $292 \times 10^3$ & & & \\
\hline
\multirow{9}{*}{\makecell{\textbf{TS-based}\\ \textbf{Multi-CLP} \\ {(ours)}}} & \multirow{3}{*}{$CLP_1$} &  \multirow{3}{*}{$5$} & \multirow{3}{*}{$24$} & $3a$ & $633 \times 10^3$ & \multirow{8}{*}{$15.32$} & \multirow{8}{*}{$2,240$} & \multirow{8}{*}{$648$} \\
\cdashline{5-6}
    & &  & & $4a$ & $475 \times 10^3$ & & & \\
\cdashline{5-6}
  & &  & & $5a$ & $356 \times 10^3$ & & & \\
 \cdashline{2-6}
 & $CLP_2$ & $3$ & $12$ & $1a$ & $1,464 \times 10^3$ & & & \\
 \cdashline{2-6}
   & $CLP_3$ & $3$ & $12$ & $1b$ & $1,464 \times 10^3$ & & & \\
  \cdashline{2-6}
    & \multirow{4}{*}{$CLP_4$} & \multirow{4}{*}{$8$} & \multirow{4}{*}{$32$} & $2a \;\; , \;\; 2b$ & $\mathbf{875 \times 10^3}$ & & & \\
    \cdashline{5-6}
    & &  & & $3b$ & $\mathbf{292 \times 10^3}$ & & & \\
\cdashline{5-6}
    & &  & & $4b$ & $\mathbf{219 \times 10^3}$ & & & \\
    \cdashline{5-6}
    & &  & & $5b$ & $\mathbf{146 \times 10^3}$ & & & \\
\hline
\end{tabular}
\end{center}
\end{table*}

Next, we use the $\textnormal{SA-based}$ $\textnormal{Multi-CLP}$ and $\textnormal{TS-based}$ $\textnormal{Multi-CLP}$ approaches to accelerate AlexNet architecture on 485T FPGA. The obtained designs are then compared with the state-of-the-art $\textnormal{Single-CLP}$~\cite{zhang2015optimizing} design and $\textnormal{Multi-CLP}$~\cite{shen2017maximizing} design, as shown in Table~\ref{tab:AlexNet_fp32_485T}. The table shows the number of CLPs adopted by each technique and the parallelism factors for each CLP. Additionally, it shows the assignment of CONV layers to CLPs. The number of cycles each CLP spends in processing its layers is also provided. Note that when layer assignment column contains more than one layer in a row, the cycle count provided is the total cycles needed to process all of those layers. 

The table also shows the execution time of an image's CONV layer, which represents the time interval in which the system can receive a new image for processing as discussed in Section~\ref{sec:multi_clp_accelerator_design}. The execution time is calculated by dividing design cycles count, illustrated in Equation~\eqref{eq:design_cycles}, over design operating frequency. Specifically, the cycle count for each CLP is the number of cycles required to execute all of its assigned layers. On the other hand, CLPs in $\textnormal{Multi-CLP}$ design work concurrently. Therefore, the overall cycle count for such a design is the maximum cycle count for its CLPs. The last two columns in Table~\ref{tab:AlexNet_fp32_485T} present the number of DSP slices and BRAMs utilized for each technique.

\begin{table*}[t!]
\renewcommand{\arraystretch}{1.4}
\begin{center}
\captionsetup{justification=centering}
\caption{Single-CLP and Multi-CLP accelerators for AlexNet on 690T FPGA using 32-bit floating-point data.}
\label{tab:AlexNet_fp32_690T}
\footnotesize
\begin{tabular}{| l || c || >{\centering}p{0.52in} | >{\centering}p{0.52in} || c || c || c || c || c |}
\hline
\multirow{2}{*}{\textbf{Technique}} & \multirow{2}{*}{\textbf{CLP No.}} & \multicolumn{2}{ c ||}{\multirow{1}{*}{\textbf{\textbf{Unrolling Factors}}}} & \multicolumn{1}{ c ||}{\multirow{2}{*}{\makecell{\textbf{Layer} \\ \textbf{Mapping}}}} & \multicolumn{1}{ c ||}{\multirow{2}{*}{\textbf{Cycles}}} & \multicolumn{1}{ c ||}{\multirow{2}{*}{\makecell{\textbf{Execution} \\ \textbf{Time (ms)}}}} & \multicolumn{1}{ c ||}{\multirow{2}{*}{\textbf{DSPs}}} & \multicolumn{1}{ c |}{\multirow{2}{*}{\textbf{BRAMs}}} \\
\cline{3-4}
& & $\mathbfit{T_n}$ & $\mathbfit{T_m}$ & & & & & \\
\hline
\hline
\multirow{5}{*}{\makecell{\textbf{Single-CLP} \\ \textbf{\cite{zhang2015optimizing}}}} & \multirow{5}{*}{$CLP_1$} & \multirow{5}{*}{$9$} & \multirow{5}{*}{$64$} & $1a \;\; , \;\; 1b$ & $\mathbf{732 \times 10^3}$ & \multirow{5}{*}{$17.69$} & \multirow{5}{*}{$2,880$} & \multirow{5}{*}{$758$} \\
\cdashline{5-6}
 & &  & & $2a \;\; , \;\; 2b$ & $\mathbf{437 \times 10^3}$ & & & \\
 \cdashline{5-6}
  & &  & & $3a \;\; , \;\; 3b$ & $\mathbf{265 \times 10^3}$ & & & \\
  \cdashline{5-6}
   & &  & & $4a\;\; , \;\; 4b$ & $\mathbf{201 \times 10^3}$ & & & \\
   \cdashline{5-6}
    & &  & & $5a \;\; , \;\; 5b$ & $\mathbf{134 \times 10^3}$ & & & \\
\hline
\multirow{6}{*}{\makecell{\textbf{Multi-CLP} \\ \textbf{\cite{shen2017maximizing}}}} & $CLP_1$ & $1$ & $64$ & $5a \;\; , \;\; 5b$ & $1,168 \times 10^3$ & \multirow{6}{*}{$11.68$} & \multirow{6}{*}{$2,880$} & \multirow{6}{*}{$1,238$} \\
\cdashline{2-6}
 & $CLP_2$ & $1$ & $96$ & $4a \;\; , \;\; 4b$ & $\mathbf{1,168 \times 10^3}$ & & & \\
\cdashline{2-6}
  & $CLP_3$ &  $2$ & $64$ & $3a \;\; , \;\; 3b$ & $1,168 \times 10^3$ & & & \\
 \cdashline{2-6}
   & $CLP_4$ & $1$ & $48$ & $1a$ & $1,098 \times 10^3$ & & & \\
   \cdashline{2-6}
    & $CLP_5$ & $1$ & $48$ & $1b$ & $1,098 \times 10^3$ & & & \\
   \cdashline{2-6}
    & $CLP_6$ & $3$ & $64$ & $2a \;\; , \;\; 2b$ & $1,166 \times 10^3$ & & & \\
\hline
\multirow{8}{*}{\makecell{\textbf{SA-based}\\ \textbf{Multi-CLP} \\ {(ours)}}} & $CLP_1$ & $3$ & $16$ & $1a$ & $1,098 \times 10^3$ & \multirow{8}{*}{$11.68$} & \multirow{8}{*}{$2,880$} & \multirow{8}{*}{$1,344$} \\
\cdashline{2-6}
 & $CLP_2$ & $3$ & $16$ & $1b$ & $1,098 \times 10^3$ & & & \\
\cdashline{2-6}
  & $CLP_3$ &  $12$ & $8$ & $2a$ & $1,166 \times 10^3$ & & & \\
 \cdashline{2-6}
   & $CLP_4$ & $6$ & $16$ & $2b$ & $\mathbf{1,166 \times 10^3}$ & & & \\
   \cdashline{2-6}
    & \multirow{3}{*}{$CLP_5$} & \multirow{3}{*}{$16$} & \multirow{3}{*}{$16$} & $3a \;\; , \;\; 3b$ & $584 \times 10^3$ & & & \\
    \cdashline{5-6}
    &  &  &  & $4a \;\; , \;\; 4b$ & $438 \times 10^3$ & & & \\
    \cdashline{5-6}
    &  &  &  & $5a$ & $146 \times 10^3$ & & & \\
   \cdashline{2-6}
    & $CLP_6$ & $8$ & $4$ & $5b$ & $1,168 \times 10^3$ & & & \\
\hline
\multirow{9}{*}{\makecell{\textbf{TS-based}\\ \textbf{Multi-CLP} \\ {(ours)}}} & \multirow{3}{*}{$CLP_1$} &  \multirow{3}{*}{$6$} & \multirow{3}{*}{$32$} & $2a$ & $583 \times 10^3$ & \multirow{9}{*}{$11.81$} & \multirow{9}{*}{$2,880$} & \multirow{9}{*}{$1,348$} \\
\cdashline{5-6}
    & &  & & $3b$ & $392 \times 10^3$ & & & \\
\cdashline{5-6}
  & &  & & $5a$ & $195 \times 10^3$ & & & \\
 \cdashline{2-6}
 & \multirow{3}{*}{$CLP_2$} &  \multirow{3}{*}{$3$} & \multirow{3}{*}{$48$} & $1a$ & $\mathbf{366 \times 10^3}$ &  &  &  \\
\cdashline{5-6}
    & &  & & $3a$ & $\mathbf{523 \times 10^3}$ & & & \\
\cdashline{5-6}
  & &  & & $5b$ & $\mathbf{292 \times 10^3}$ & & & \\
 \cdashline{2-6}
   & $CLP_3$ & $3$ & $16$ & $1b$ & $1,098 \times 10^3$ & & & \\
  \cdashline{2-6}
    & \multirow{2}{*}{$CLP_4$} & \multirow{2}{*}{$6$} & \multirow{2}{*}{$32$} & $2b$ & $583 \times 10^3$ & & & \\
    \cdashline{5-6}
    & &  & & $4a \;\; , \;\; 4b$ & $584 \times 10^3$ & & & \\
\hline
\end{tabular}
\end{center}
\vskip 10pt
\end{table*}

As evidenced by the results, the number of cycles required in the case of a single CLP implementation is almost $1.3\times$ more than that of the multiple CLPs implementations. The results also show that $\textnormal{SA-based}$ $\textnormal{Multi-CLP}$ and $\textnormal{TS-based}$ $\textnormal{Multi-CLP}$ were able to find implementation configurations with a fewer number of cycles (less execution time) compared to $\textnormal{Multi-CLP}$ even though all designs use the same number of CLPs and DSP slices. This demonstrates the effectiveness of the proposed $\textnormal{SA-based}$ $\textnormal{Multi-CLP}$ and $\textnormal{TS-based}$ $\textnormal{Multi-CLP}$ in balancing the computational resources and workloads assigned to each CLPs such that each CLP can be kept busy most of the time. Although the implementation configurations obtained using SA and TS metaheuristic algorithms show small improvements, the results indicate  the importance of using metaheuristics in finding different configurations for a given CNN using a systematic methodology.

Table~\ref{tab:AlexNet_fp32_690T} shows similar experiments performed for AlexNet on 690T FPGA. The number of cycles required in the case of a single CLP implementation is about $1.5\times$ more than that of multiple CLPs implementations. For the implementation configuration obtained using $\textnormal{SA-based}$ $\textnormal{Multi-CLP}$, we got the same number of cycles as obtained by $\textnormal{Multi-CLP}$. Although both techniques result in designs with same number of cycles and CLPs, these designs have completely different configurations for their adopted CLPs. On the other hand, implementation configuration obtained by $\textnormal{TS-based}$ $\textnormal{Multi-CLP}$ has used only $4$ CLPs and results in a higher execution time.

\begin{table*}[t!]
\renewcommand{\arraystretch}{1.3}
\begin{center}
\captionsetup{justification=centering}
\caption{AlexNet accelerators performance on 485T and 690T FPGAs using 32-bit floating-point data.}
\label{tab:AlexNet_overall_performance}
\footnotesize
\begin{tabular}{| l || >{\centering}p{0.52in} | >{\centering}p{0.52in} | >{\centering}p{0.52in} | c || >{\centering}p{0.52in} | >{\centering}p{0.52in} | >{\centering}p{0.52in} | c |}
\hline
\multicolumn{1}{| l ||}{\multirow{3}{*}{\textbf{Technique}}} & \multicolumn{4}{ c ||}{\multirow{1}{*}{\textbf{Virtex-7 VX485T}}}  & \multicolumn{4}{ c |}{\multirow{1}{*}{\textbf{Virtex-7 VX690T}}}\\
\cline{2-9}
 & \multicolumn{1}{ c |}{\multirow{2}{*}{\makecell{\textbf{BW} \\ \textbf{(GB/s)}}}} & \multicolumn{1}{ c |}{\multirow{2}{*}{\makecell{\textbf{Arith.} \\ \textbf{Util.}}}}  & \multicolumn{1}{ c |}{\multirow{2}{*}{\makecell{\textbf{Thr.} \\ \textbf{(img/s)}}}}  & \multicolumn{1}{ c ||}{\multirow{2}{*}{\makecell{\textbf{Perf.} \\ \!\!\textbf{(GFLOPs/s)}\!\!}}} & \multicolumn{1}{ c |}{\multirow{2}{*}{\makecell{\textbf{BW} \\ \textbf{(GB/s)}}}} & \multicolumn{1}{ c |}{\multirow{2}{*}{\makecell{\textbf{Arith.} \\ \textbf{Util.}}}}  & \multicolumn{1}{ c |}{\multirow{2}{*}{\makecell{\textbf{Thr.} \\ \textbf{(img/s)}}}}  & \multicolumn{1}{ c |}{\multirow{2}{*}{\makecell{\textbf{Perf.} \\ \!\!\textbf{(GFLOPs/s)}\!\!}}} \\
& & & &  & & & & \\
\hline
\hline
\multirow{2}{*}{\makecell{\textbf{Single-CLP} \\ \textbf{\cite{zhang2015optimizing}}}} & \multirow{2}{*}{$1.40$} & \multirow{2}{*}{$74.1$\%} & \multirow{2}{*}{$48.85$} &  \multirow{2}{*}{$65.05$}  & \multirow{2}{*}{$1.78$} & \multirow{2}{*}{$64.0$\%} & \multirow{2}{*}{$55.40$} &  \multirow{2}{*}{$73.77$} \\
& & & &  & & & & \\
\hline
\multirow{2}{*}{\makecell{\textbf{Multi-CLP} \\ \textbf{\cite{shen2017maximizing}}}} & \multirow{2}{*}{$1.38$} & \multirow{2}{*}{$95.6$\%} & \multirow{2}{*}{$63.98$} &  \multirow{2}{*}{$85.20$}  & \multirow{2}{*}{$1.49$} & \multirow{2}{*}{$98.1$\%} & \multirow{2}{*}{$85.55$} &  \multirow{2}{*}{$113.92$} \\
& & & &  & & & & \\
\hline
\multirow{3}{*}{\makecell{\textbf{SA-based}\\ \textbf{Multi-CLP} \\ {(ours)}}} & \multirow{3}{*}{$1.42$} & \multirow{3}{*}{$97.4$\%} & \multirow{3}{*}{$65.32$} &  \multirow{3}{*}{$86.98$} & \multirow{3}{*}{$1.35$} & \multirow{3}{*}{$98.1$\%} & \multirow{3}{*}{$85.55$} &  \multirow{3}{*}{$113.92$} \\
& & & &  & & & & \\
& & & &  & & & & \\
\hline
\multirow{3}{*}{\makecell{\textbf{TS-based}\\ \textbf{Multi-CLP} \\ {(ours)}}} & \multirow{3}{*}{$1.33$} & \multirow{3}{*}{$95.7$\%} & \multirow{3}{*}{$65.27$} &  \multirow{3}{*}{$86.91$} & \multirow{3}{*}{$1.30$} & \multirow{3}{*}{$98.1$\%} & \multirow{3}{*}{$84.67$} &  \multirow{3}{*}{$112.75$} \\
& & & &  & & & & \\
& & & &  & & & & \\
\hline
\end{tabular}
\end{center}
\end{table*}

\begin{table*}[th!]
\renewcommand{\arraystretch}{1.3}
\begin{center}
\captionsetup{justification=centering}
\caption{Single-/Multi-CLP accelerators for SqueezeNet on 485T and 690T FPGAs using 16-bit fixed-point data.}
\label{tab:SqueezeNet_fxp16_485t_690T}
\footnotesize
\begin{tabular}{| l || >{\centering}p{0.35in} | >{\centering}p{0.25in} | >{\centering}p{0.25in} | >{\centering}p{0.65in} | c || >{\centering}p{0.35in} | >{\centering}p{0.25in} | >{\centering}p{0.25in} | >{\centering}p{0.65in} | c |}
\hline
\multicolumn{1}{| l ||}{\multirow{3}{*}{\textbf{Technique}}} & \multicolumn{5}{ c ||}{\multirow{1}{*}{\textbf{Virtex-7 VX485T}}}  & \multicolumn{5}{ c |}{\multirow{1}{*}{\textbf{Virtex-7 VX690T}}}\\
\cline{2-11}
 & \multicolumn{1}{ c |}{\multirow{2}{*}{\makecell{\textbf{CLP} \\ \textbf{No.}}}} & \multicolumn{2}{ c |}{\multirow{1}{*}{\textbf{Unrolling}}}  & \multicolumn{1}{ c |}{\multirow{2}{*}{\makecell{\textbf{Layer} \\ \textbf{Mapping}}}}  & \multicolumn{1}{ c ||}{\multirow{2}{*}{\textbf{Cycles}}} & \multicolumn{1}{ c |}{\multirow{2}{*}{\makecell{\textbf{CLP} \\ \textbf{No.}}}} & \multicolumn{2}{ c |}{\multirow{1}{*}{\textbf{Unrolling}}}  & \multicolumn{1}{ c |}{\multirow{2}{*}{\makecell{\textbf{Layer} \\ \textbf{Mapping}}}}  & \multicolumn{1}{ c |}{\multirow{2}{*}{\textbf{Cycles}}} \\
 \cline{3-4}
 \cline{8-9}
& & $\mathbfit{T_n}$ & $\mathbfit{T_m}$ &  & & & $\mathbfit{T_n}$ & $\mathbfit{T_m}$ & & \\
\hline
\hline
\multirow{2}{*}{\makecell{\textbf{Single-CLP} \\ \textbf{\cite{zhang2015optimizing}}}} & \multirow{2}{*}{$CLP_1$} & \multirow{2}{*}{$32$} & \multirow{2}{*}{$68$} & \multirow{2}{*}{$1 - 26$} & \multirow{2}{*}{$\mathbf{349 \times 10^3}$} & \multirow{2}{*}{$CLP_1$} & \multirow{2}{*}{$32$} & \multirow{2}{*}{$87$} & \multirow{2}{*}{$1 - 26$} & \multirow{2}{*}{$\mathbf{331 \times 10^3}$} \\
& & & &  & & & & & & \\
\hline
\multirow{6}{*}{\makecell{\textbf{Multi-CLP} \\ \textbf{\cite{shen2017maximizing}}}} & $CLP_1$ & $6$ & $16$ & $2, 3, 6, 5$ & $179 \times 10^3$ & $CLP_1$ & $8$ & $16$ & $2, 6, 3, 5$ & $125 \times 10^3$ \\
\cdashline{2-11}
 & $CLP_2$ & $3$ & $64$ & $1, 8, 9, 12$ & $183 \times 10^3$ & $CLP_2$ & $3$ & $64$ & $1$ & ${115 \times 10^3}$ \\
 \cdashline{2-11}
 & $CLP_3$ & $4$ & $64$ & $all \; others$ & ${165 \times 10^3}$ & $CLP_3$ & $11$ & $32$ & $all \; others$ & ${133 \times 10^3}$ \\
 \cdashline{2-11}
 & $CLP_4$ & $8$ & $64$ & $7, 4, 16, 19$ & ${176 \times 10^3}$ & $CLP_4$ & $8$ & $64$ & $7, 4, 16$ & $\mathbf{145 \times 10^3}$ \\
 \cdashline{2-11}
 & $CLP_5$ & $8$ & $128$ & $26, 22, 25, 13$ & $\mathbf{185 \times 10^3}$ & $CLP_5$ & $5$ & $256$ & $19, 26, 22, 25$ & ${144 \times 10^3}$ \\
 \cdashline{2-11}
 & $CLP_6$ & $16$ & $10$ & $10$ & ${183 \times 10^3}$ & $CLP_6$ & $16$ & $26$ & $13, 10$ & ${141 \times 10^3}$ \\
\hline
\multirow{10}{*}{\makecell{\textbf{SA-based} \\ \textbf{Multi-CLP} \\ {(ours)}}} & $CLP_1$ & $2$ & $7$ & $14$ & ${176 \times 10^3}$ & $CLP_1$ & $64$ & $16$ & $5, 23, 25, 26$ & $\mathbf{139.5 \times 10^3}$ \\
\cdashline{2-11}
 & $CLP_2$ & $16$ & $32$ & $7, 13, 20, 25$ & ${179 \times 10^3}$ & $CLP_2$ & $3$ & $64$ & $1, 12$ & ${132 \times 10^3}$ \\
\cdashline{2-11}
 & $CLP_3$ & $8$ & $16$ & $2, 16, 17$ & ${180 \times 10^3}$ & $CLP_3$ & $32$ & $8$ & $11, 13$ & ${138 \times 10^3}$ \\
\cdashline{2-11}
 & $CLP_4$ & $13$ & $19$ & $5, 10$ & ${148 \times 10^3}$ & $CLP_4$ & $8$ & $22$ & $2, 16, 24$ & ${139 \times 10^3}$ \\
\cdashline{2-11}
 & $CLP_5$ & $43$ & $13$ & $26$ & $\mathbf{181 \times 10^3}$ & $CLP_5$ & $8$ & $32$ & $3, 4, 9$ & ${138 \times 10^3}$ \\
\cdashline{2-11}
 & $CLP_6$ & $11$ & $4$ & $3, 8$ & ${176 \times 10^3}$ & $CLP_6$ & $8$ & $32$ & $10, 18, 20$ & ${139 \times 10^3}$ \\
\cdashline{2-11}
 & $CLP_7$ & $3$ & $64$ & $1, 6, 23, 15$ & ${177 \times 10^3}$ & $CLP_7$ & $8$ & $43$ & $19, 22$ & ${138 \times 10^3}$ \\
\cdashline{2-11}
 & $CLP_8$ & $16$ & $13$ & $4, 18, 24, 12$ & ${181 \times 10^3}$ & $CLP_8$ & $8$ & $32$ & $7, 6, 14$ & ${138 \times 10^3}$ \\
\cdashline{2-11}
 & $CLP_9$ & $8$ & $26$ & $9, 21, 22$ & ${172 \times 10^3}$ & $CLP_9$ & $7$ & $8$ & $8, 15$ & ${93 \times 10^3}$ \\
\cdashline{2-11}
 & $CLP_{10}$ & $16$ & $8$ & $11, 19$ & ${177 \times 10^3}$ & $CLP_{10}$ & $14$ & $4$ & $17, 21$ & ${129 \times 10^3}$ \\
\hline
\multirow{11}{*}{\makecell{\textbf{TS-based} \\ \textbf{Multi-CLP} \\ {(ours)}}}  & $CLP_1$ & $8$ & $26$ & $3, 22, 6$ & ${179 \times 10^3}$ & $CLP_1$ & $3$ & $66$ & $21, 1$ & ${132 \times 10^3}$ \\
\cdashline{2-11}
 & $CLP_2$ & $8$ & $22$ & $7$ & ${170 \times 10^3}$ & $CLP_2$ & $12$ & $4$ & $5$ & ${138 \times 10^3}$ \\
\cdashline{2-11}
 & $CLP_3$ & $11$ & $53$ & $26, 9$ & ${182 \times 10^3}$ & $CLP_3$ & $16$ & $32$ & $6, 11, 7, 20, 25$ & ${141 \times 10^3}$ \\
\cdashline{2-11}
 & $CLP_4$ & $16$ & $10$ & $10$ & ${176 \times 10^3}$ & $CLP_4$ & $43$ & $1$ & $17$ & ${85 \times 10^3}$ \\
\cdashline{2-11}
 & $CLP_5$ & $3$ & $66$ & $1, 14, 21, 23$ & ${183 \times 10^3}$ & $CLP_5$ & $16$ & $16$ & $2, 4, 14$ & ${135 \times 10^3}$ \\
\cdashline{2-11}
 & $CLP_6$ & $16$ & $22$ & $4, 18, 25$ & ${175 \times 10^3}$ & $CLP_6$ & $16$ & $2$ & $3$ & ${100 \times 10^3}$ \\
\cdashline{2-11}
 & $CLP_7$ & $10$ & $8$ & $8, 11, 20$ & $\mathbf{183 \times 10^3}$ & $CLP_7$ & $16$ & $32$ & $all \; others$ & ${139 \times 10^3}$ \\
\cdashline{2-11}
 & $CLP_8$ & $11$ & $19$ & $2, 13, 24$ & ${183 \times 10^3}$ & $CLP_8$ & $8$ & $22$ & $19, 12, 24$ & ${133 \times 10^3}$ \\
\cdashline{2-11}
 & $CLP_9$ & $6$ & $32$ & $15, 16, 19$ & ${179 \times 10^3}$ & $CLP_9$ & $3$ & $41$ & $16$ & $\mathbf{141 \times 10^3}$ \\
\cdashline{2-11}
 & \multirow{2}{*}{$CLP_{10}$} & \multirow{2}{*}{$10$} & \multirow{2}{*}{$8$} & \multirow{2}{*}{$5, 12, 17$} & \multirow{2}{*}{${178 \times 10^3}$} & $CLP_{10}$ & $8$ & $26$ & $10$ & ${141 \times 10^3}$ \\
 \cdashline{7-11}
 &  &  & &  &  & $CLP_{11}$ & $47$ & $16$ & $8, 26$ & ${140.5 \times 10^3}$ \\
\hline
\end{tabular}
\end{center}
\end{table*}

In Table~\ref{tab:AlexNet_overall_performance}, we summarize the results presented in Tables~\ref{tab:AlexNet_fp32_485T}~and~\ref{tab:AlexNet_fp32_690T} to get an insight into the required bandwidth, arithmetic unit utilization, CONV layers throughput, and performance. We can see that $\textnormal{SA-/TS-based}$ $\textnormal{Multi-CLP}$ provides $1.34 \times$ and $1.54 \times$ more throughput improvement than $\textnormal{Single-CLP}$ design on 485T and 690T FPGAs, respectively. This improvement comes from its ability to make better use of the available arithmetic units. Specifically, $\textnormal{Single-CLP}$ can provide a useful work to the multipliers and adders only $74$\% and $64$\% of the time on 485T and 690T FPGAs, respectively, whereas $\textnormal{SA-/TS-based}$ $\textnormal{Multi-CLP}$ improves the arithmetic units utilization to about $96$\% and $98$\% on 485T and 690T FPGAs, respectively. Note that the arithmetic utilization is computed as discussed in Equation~\eqref{eq:utilization}. Furthermore, $\textnormal{SA-/TS-based}$ $\textnormal{Multi-CLP}$ improves CONV layer throughput of $\textnormal{Multi-CLP}$ design by $1.02 \times$ on 485T.

On the other hand, there is a trade-off between off-chip memory bandwidth and on-chip buffer size. Using large buffers saves off-chip memory bandwidth, whereas adopting small buffers results in a high bandwidth requirement. Here, we must emphasize that the proposed optimization framework gives a higher cost to bandwidth requirement than that given to buffer size. In other words, when more than one feasible design with the same low number of execution cycles is encountered, the design with minimum bandwidth requirement is chosen as optimal. We have adopted this strategy to minimize power expenses since on-chip memory modules consume less power than off-chip memory modules~\cite{garcia2019optimized}.

The results for single CLP and multiple CLPs accelerators for SqueezeNet~$1.1$ on 485T and 690T FPGAs with $16$-bit fixed-point data are shown in Table~\ref{tab:SqueezeNet_fxp16_485t_690T}. For SqueezeNet~$1.1$ accelerators on 485T FPGA, the single CLP implementation requires about $1.9\times$ more cycles than the multiple CLPs implementation. Even more importantly, though, the results show that as the number of CONV layers increases, the search space also increases, making it more challenging to obtain an optimal multiple CLPs design using conventional iterative algorithms. Specifically, $\textnormal{SA-based}$ $\textnormal{Multi-CLP}$ processes images in a $1.04$~ms and a $1.43$~ms shorter time than $\textnormal{Multi-CLP}$ on 485T and 690T FPGAs, respectively.  Additionally, although $\textnormal{TS-based}$ $\textnormal{Multi-CLP}$ produced a solution that requires more cycles than that of $\textnormal{SA-based}$ $\textnormal{Multi-CLP}$, it processes images in a shorter time and with fewer DSP slices than $\textnormal{Multi-CLP}$.

On the other hand, when comparing the performance of $\textnormal{SA-based}$ $\textnormal{Multi-CLP}$ in accelerating SqueezeNet~$1.1$ on VC707 and VC709 boards, we can notice that it used the available resources to the maximum extent to improve the network throughput. Specifically, with a $1.29\times$ increase in computational resources in VC709 compared to VC707, $\textnormal{SA-based}$ $\textnormal{Multi-CLP}$ increased SqueezeNet~$1.1$ throughput by $1.30\times$, which demonstrates the scalability of the proposed accelerator.

In Tables \ref{tab:vggnet_fxp16_690T} and
\ref{tab:GoogLeNet_fxp16_690T}, we show the configurations of single CLP and multiple CLPs accelerators designed for VGGNet and GoogLeNet on 690T FPGA with $16$-bit fixed-point representation. Note that no results have been reported in $\textnormal{Single-CLP}$~\cite{zhang2015optimizing} and $\textnormal{Multi-CLP}$~\cite{shen2017maximizing} accelerators for these two large CNNs. Therefore, with the sake of brevity, we only present the results for $\textnormal{SA-based}$ Single-/Multi-CLP as it turns out from previous experiments that SA algorithm provides more optimized solutions than TS algorithm in this specific problem.

For VGGNet, $\textnormal{SA-based}$ $\textnormal{Multi-CLP}$ reduces the total cycles compared to $\textnormal{SA-based}$ $\textnormal{Single-CLP}$ by about ${676 \times 10^3}$ cycles. Here, we see a significant difference compared to the results from AlexNet, which is due to the variance in the dimensions between CONV layers in VGGNet, where, for example, the first CONV layer has $\langle N, M \rangle$ as $\langle 3, 64 \rangle$, and those for last CONV layer is $\langle 512, 512 \rangle$. Hence, using a single CLP  results in a large  DSP resource under-utilization during the processing of some layers. For GoogLeNet, it requires around $2\times$ more cycles  for the single CLP implementation than that of multiple CLPs. For large CNNs, such as GoogLeNet which has $57$ CONV layers, it becomes more challenging to find an optimized multiple CLPs configuration without using metaheuristics algorithms.

Additionally, we modeled the behavior of multiple CLPs design in Verilog and synthesized it using Xilinx Vivado Design Suite targeting the Xilinx Virtex-7 485T FPGA. We kept all synthesis properties to their default. Table~\ref{tab:AlexNet_overall_implementation_performance} shows a summary of the implementation and performance results for the proposed $\textnormal{SA-based}$ $\textnormal{Multi-CLP}$ approach and compares them with those reported in $\textnormal{Single-CLP}$~\cite{zhang2015optimizing}, $\textnormal{GA-based}$ $\textnormal{Single-CLP}$~\cite{suda2016throughput}, and $\textnormal{Multi-CLP}$~\cite{shen2017maximizing} approaches when they are used to accelerate AlexNet architecture. For each technique, the table provides the FPGA platform used for implementation, operating frequency, precision adopted to represent FMs and weights, design entry employed to describe the accelerator, hardware resource usage after placement and routing in terms of DSPs, BRAMs, FFs, and LUTs, and power efficiency (a.k.a., performance per Watt).

\begin{table}[t!]
\renewcommand{\arraystretch}{1.3}
\begin{center}
\captionsetup{justification=centering}
\caption{SA-based Single-/Multi-CLP accelerators for VGGNet on 690T FPGA using 16-bit fixed-point data.}
\label{tab:vggnet_fxp16_690T}
\footnotesize
\begin{tabular}{| l | >{\centering}p{0.3in} | >{\centering}p{0.2in} | >{\centering}p{0.2in} | c | c |}
\hline
 \multicolumn{1}{| l |}{\multirow{2}{*}{\textbf{Technique}}} & \multicolumn{1}{ c |}{\multirow{2}{*}{\makecell{\textbf{CLP} \\ \textbf{No.}}}} & \multicolumn{2}{ c |}{\multirow{1}{*}{\textbf{Unrolling}}}  & \multicolumn{1}{ c |}{\multirow{2}{*}{\makecell{\textbf{Layer} \\ \textbf{Mapping}}}}  & \multicolumn{1}{ c |}{\multirow{2}{*}{\makecell{\textbf{Cycles} \\ \textbf{($\times 10^3$)}}}} \\
 \cline{3-4}
& & $\mathbfit{T_n}$ & $\mathbfit{T_m}$ &  & \\
 \hline
\hline
\multirow{3}{*}{\makecell{\textbf{SA-based} \\ \textbf{Single-CLP} \\ {(ours)}}} & \multirow{3}{*}{$CLP_1$} & \multirow{3}{*}{$44$} & \multirow{3}{*}{$65$} & \multirow{3}{*}{$1 - 13$} & \multirow{3}{*}{$\mathbf{6,631}$} \\
& & & &  & \\
& & & &  & \\
\hline
\multirow{4}{*}{\makecell{\textbf{SA-based} \\ \textbf{Multi-CLP} \\ {(ours)}}} & $CLP_1$ & $8$ & $64$ & $1, 8, 4$ & $\mathbf{5,955}$ \\
\cdashline{2-6}
 & $CLP_2$ & $32$ & $19$ & $3, 7, 11$ & ${5,503}$ \\
\cdashline{2-6}
 & $CLP_3$ & $48$ & $2$ & $13$ & ${4,976}$ \\
\cdashline{2-6}
 & $CLP_4$ & $64$ & $26$ & $all \; others$ & ${5,841}$ \\
\hline
\end{tabular}
\end{center}
\end{table}

When comparing resource usage estimates with resource usage from Vivado’s implementation report for $\textnormal{SA-based}$ $\textnormal{Multi-CLP}$, one can note that the model underestimated the DSP slices count by $53$ DSP slices per CLP, on average. The reason is that Vivado’s utilization report for $\textnormal{SA-based}$ $\textnormal{Multi-CLP}$ design accounts not only for DSPs used in CLP’s computational engine, but also for those used in the control logic, address calculations, and loop indexing as well. When considering only those DSPs used to implement the computational engines for $\textnormal{SA-based}$ $\textnormal{Multi-CLP}$, we found that the estimates match those reported by Vivado utilization report.

\begin{table}[t!]
\renewcommand{\arraystretch}{1.3}
\begin{center}
\captionsetup{justification=centering}
\caption{SA-based Single-/Multi-CLP accelerators for GoogLeNet on 690T FPGA using 16-bit fixed-point.}
\label{tab:GoogLeNet_fxp16_690T}
\footnotesize
\begin{tabular}{| l | >{\centering}p{0.3in} | >{\centering}p{0.2in} | >{\centering}p{0.2in} | c | c |}
\hline
 \multicolumn{1}{| l |}{\multirow{2}{*}{\textbf{Technique}}} & \multicolumn{1}{ c |}{\multirow{2}{*}{\makecell{\textbf{CLP} \\ \textbf{No.}}}} & \multicolumn{2}{ c |}{\multirow{1}{*}{\textbf{Unrolling}}}  & \multicolumn{1}{ c |}{\multirow{2}{*}{\makecell{\textbf{Layer} \\ \textbf{Mapping}}}}  & \multicolumn{1}{ c |}{\multirow{2}{*}{\makecell{\textbf{Cycles} \\ \textbf{($\times 10^3$)}}}} \\
 \cline{3-4}
& & $\mathbfit{T_n}$ & $\mathbfit{T_m}$ &  & \\
 \hline
\hline
\multirow{3}{*}{\makecell{\textbf{SA-based} \\ \textbf{Single-CLP} \\ {(ours)}}} & \multirow{3}{*}{$CLP_1$} & \multirow{3}{*}{$45$} & \multirow{3}{*}{$64$} & \multirow{3}{*}{$1 - 57$} & \multirow{3}{*}{$\mathbf{1,330}$} \\
& & & &  & \\
& & & &  & \\
\hline
\multirow{27}{*}{\makecell{\textbf{SA-based} \\ \textbf{Multi-CLP} \\ {(ours)}}} & 
\multirow{2}{*}{$CLP_1$} & \multirow{2}{*}{$51$} & \multirow{2}{*}{$2$} & $11, 23, 31,$ & \multirow{2}{*}{${620}$} \\
&  & &  & $43, 28$ &  \\
\cdashline{2-6}
 & $CLP_2$ & $1$ & $21$ & $32$ & ${470}$ \\
\cdashline{2-6}
 & $CLP_3$ & $10$ & $64$ & $3, 19, 53$ & ${636}$ \\
\cdashline{2-6}
 & $CLP_4$ & $78$ & $1$ & $17, 21, 40$ & ${571}$ \\
\cdashline{2-6}
 & $CLP_5$ & $26$ & $2$ & $8, 15$ & ${564}$ \\
\cdashline{2-6}
 & $CLP_6$ & $8$ & $20$ & $36, 16$ & ${594}$ \\
\cdashline{2-6}
 & $CLP_7$ & $4$ & $13$ & $54$ & ${635}$ \\
\cdashline{2-6}
 & $CLP_8$ & $2$ & $40$ & $10, 45$ & ${608}$ \\
\cdashline{2-6}
 & \multirow{2}{*}{$CLP_9$} & \multirow{2}{*}{$18$} & \multirow{2}{*}{$3$} & $37, 38, 51,$ & \multirow{2}{*}{${634}$} \\
 & & & & $56, 57$ & \\
\cdashline{2-6}
 & $CLP_{10}$ & $7$ & $13$ & $18, 34, 50$ & ${587}$ \\
\cdashline{2-6}
 & $CLP_{11}$ & $8$ & $13$ & $30, 47$ & ${631}$ \\
\cdashline{2-6}
 & $CLP_{12}$ & $9$ & $32$ & $12$ & ${635}$ \\
\cdashline{2-6}
 & $CLP_{13}$ & $14$ & $1$ & $27$ & ${464}$ \\
 \cdashline{2-6}
 & $CLP_{14}$ & $3$ & $66$ & $1$ & $\mathbf{637}$ \\
\cdashline{2-6}
 & $CLP_{15}$ & $1$ & $11$ & $49$ & ${122}$ \\
\cdashline{2-6}
 & $CLP_{16}$ & $6$ & $11$ & $13, 39, 44$ & ${555}$ \\
\cdashline{2-6}
 & $CLP_{17}$ & $2$ & $38$ & $35, 26, 48$ & ${636}$ \\
\cdashline{2-6}
 & $CLP_{18}$ & $6$ & $27$ & $25, 42, 55$ & ${602}$ \\
\cdashline{2-6}
 & \multirow{2}{*}{$CLP_{19}$} & \multirow{2}{*}{$12$} & \multirow{2}{*}{$20$} & $9, 14,$ & \multirow{2}{*}{${575}$} \\
 & & & & $24, 46$ & \\
\cdashline{2-6}
 & $CLP_{20}$ & $4$ & $6$ & $5$ & ${602}$ \\
\cdashline{2-6}
 & \multirow{2}{*}{$CLP_{21}$} & \multirow{2}{*}{$9$} & \multirow{2}{*}{$8$} & $29, 33,$ & \multirow{2}{*}{${521}$} \\
 &  & & & $7, 52$ & \\
\cdashline{2-6}
 & $CLP_{22}$ & $98$ & $1$ & $2, 20, 41$ & ${631}$ \\
\cdashline{2-6}
 & $CLP_{23}$ & $99$ & $2$ & $4, 6, 22$ & ${596}$ \\
\hline
\end{tabular}
\end{center}
\end{table}

\begin{table*}[t!]
\renewcommand{\arraystretch}{1.3}
\begin{center}
\captionsetup{justification=centering}
\caption{Implementation and performance summary of Single-/Multi-CLP accelerators optimized for AlexNet.}
\label{tab:AlexNet_overall_implementation_performance}
\footnotesize
\resizebox{\linewidth}{!}{
\begin{tabular}{| c | c | c | c | c | >{\centering}p{0.42in} | >{\centering}p{0.42in} | >{\centering}p{0.42in} | >{\centering}p{0.42in} | c |}
\hline
 \multicolumn{1}{| c |}{\multirow{2}{*}{\textbf{Technique}}} & \multicolumn{1}{ c |}{\multirow{2}{*}{\textbf{Platform}}} & \multicolumn{1}{ c |}{\multirow{2}{*}{\textbf{\makecell{Frequency \\ (MHz)}}}} & \multicolumn{1}{ c |}{\multirow{2}{*}{\textbf{Precision}}} & \multicolumn{1}{ c |}{\multirow{2}{*}{\textbf{\makecell{Design \\ Entry}}}} & \multicolumn{4}{ c |}{\multirow{1}{*}{\textbf{Resources Utilization}}}  & \multicolumn{1}{ c |}{\multirow{2}{*}{\textbf{\makecell{Power Efficiency\\(GOPs/Watt)}}}} \\
 \cline{6-9}
& & & & & \textbf{DSP} & \textbf{BRAM}   & \textbf{FF} & \textbf{LUT} & \\
 \hline
\hline
\multirow{3}{*}{\makecell{\textbf{Single-CLP} \\ \textbf{\cite{zhang2015optimizing}}}} & \multirow{3}{*}{\makecell{Virtex-7 \\ VX485T}} & \multirow{3}{*}{$100$} & \multirow{3}{*}{\makecell{$32$-bit \\ floating-point}} & \multirow{3}{*}{C} & \multirow{3}{*}{\makecell{$2,309$ \\ ($82$\%)}} &  \multirow{3}{*}{\makecell{$689$ \\ ($34$\%)}} & \multirow{3}{*}{\makecell{$219,815$ \\ ($36$\%)}} &   \multirow{3}{*}{\makecell{$146,325$ \\ ($48$\%)}}    & \multirow{3}{*}{$3.31$} \\
 & & &   &  &  & & &      &  \\
  & & &   & &  & & &     &  \\
\hline
\multirow{3}{*}{\makecell{\textbf{GA-based} \\ \textbf{Single-CLP} \\ \textbf{\cite{suda2016throughput}}}} & \multirow{3}{*}{\makecell{Stratix-V \\ GSD8}} & \multirow{3}{*}{$120$} & \multirow{3}{*}{\makecell{($8$/$16$)-bit \\ fixed-point}} & \multirow{3}{*}{OpenCL} & \multirow{3}{*}{\makecell{$727$ \\ ($37$\%)}} &  \multirow{3}{*}{\makecell{$1,480$ \\ ($58$\%)}} & \multirow{3}{*}{N/A} &   \multirow{3}{*}{\makecell{$119,622$ \\ ($46$\%)}}    & \multirow{3}{*}{$3.79$} \\  
 & & &   & & &  &   &      &  \\
  & & &   & & & &  &     &  \\
\hline
\multirow{3}{*}{\makecell{\textbf{Multi-CLP} \\ \textbf{\cite{shen2017maximizing}}}} & \multirow{3}{*}{\makecell{Virtex-7 \\ VX485T}} & \multirow{3}{*}{$100$} & \multirow{3}{*}{\makecell{$32$-bit \\ floating-point}} & \multirow{3}{*}{C++} & \multirow{3}{*}{\makecell{$2,443$ \\ ($87$\%)}} &  \multirow{3}{*}{\makecell{$812$ \\ ($39$\%)}} & \multirow{3}{*}{\makecell{$270,991$ \\ ($45$\%)}} &   \multirow{3}{*}{\makecell{$176,876$ \\ ($58$\%)}}    & \multirow{3}{*}{$11.21$} \\
 & & &   &  & & &  &      &  \\
  & & &   & & & & &     &  \\
\hline
\multirow{3}{*}{\makecell{\textbf{SA-based} \\ \textbf{Multi-CLP} \\ {(ours)}}} & \multirow{3}{*}{\makecell{Virtex-7 \\ VX485T}} & \multirow{3}{*}{$100$} & \multirow{3}{*}{\makecell{$32$-bit \\ floating-point}} & \multirow{3}{*}{Verilog} &\multirow{3}{*}{\makecell{$2,452$ \\ ($88$\%)}} &  \multirow{3}{*}{\makecell{$644$ \\ ($31$\%)}} & \multirow{3}{*}{\makecell{$269,053$ \\ ($44$\%)}} &   \multirow{3}{*}{\makecell{$176,449$ \\ ($58$\%)}}    & \multirow{3}{*}{$12.08$} \\
 & & &   &  & & &  &      &  \\
  & & &   & &  &   & &  &  \\
\hline
\end{tabular}
}
\end{center}
\end{table*}

On the other hand, the implemented and predicted BRAM counts are perfectly matched. This is because in the implemented parametrized Verilog modules, we instantiate the exact required memory type, mode, and count using Xilinx HDL language templates. However, this does not apply to $\textnormal{Single-CLP}$, $\textnormal{GA-based}$ $\textnormal{Single-CLP}$, and $\textnormal{Multi-CLP}$ because they use high-level synthesis tools to compile C, OpenCL, and C++ codes into HDL codes, respectively. As can be seen from the results, designing with multiple CLPs makes better use of available resources than a single CLP design. Due to the dimensional mismatch issue in $\textnormal{Single-CLP}$ design, this design scheme could not take advantage of more than $82$\% of DSP slices. One can also note that the proposed $\textnormal{SA-based}$ $\textnormal{Multi-CLP}$ increases the usage of FF and LUT resources in $\textnormal{Single-CLP}$ by about $5$\% for each additional CLP. All this increase in resource usage contributes to improving the power efficiency by $3.65\times$.

Compared to $\textnormal{GA-based}$ $\textnormal{Single-CLP}$, the proposed $\textnormal{SA-based}$ $\textnormal{Multi-CLP}$ has effectively made use of the available resources resulting in a $3.19\times$ increase in power efficiency even though $\textnormal{GA-based}$ $\textnormal{Single-CLP}$ uses $8$-bit fixed-point weights and $16$-bit fixed-point FMs while our accelerator uses $32$-bit floating-point data and operates at a lower frequency. This is due to the high-level synthesis tool employed in~\cite{suda2016throughput} which constrains $T_{in}$ to be from the set $\{1, 2, 4, 8, 16\}$ and $T_{out}$ to be integer multiplicative of $T_{in}$, where $T_{out}$ and $T_{in}$ are the unified unrolling factors for the output matrix of CONV operation and the input vectors to CLP structures, respectively, which are determined by GA to minimize execution time. In contrast, this work allows unrolling factors to be whatever value yields the highest performance.

On the other hand, the usage of DSP, FF, and LUT resources for $\textnormal{Multi-CLP}$ and the proposed $\textnormal{SA-based}$ $\textnormal{Multi-CLP}$ is almost consistent. However, the additional $168$ BRAMs used in $\textnormal{Multi-CLP}$ led to a $0.4$ Watt increase in its power consumption, which also caused its power efficiency to be about $0.87$ GOPs/Watt lower than that of the proposed $\textnormal{SA-based}$ $\textnormal{Multi-CLP}$.

\section{Conclusion}
\label{sec:conclusion}

CNNs have shown great performance in a wide range of machine learning applications. However, they require significant computational power that cannot be met by general purpose processors. Nowadays, FPGAs have been found to deliver excellent performance in accelerating CNNs as they provide the best performance per Watt. Most of the existing FPGA-based accelerators use a single CLP for processing all CONV layers. However, the current trend in CNN acceleration is using multiple CLPs to make efficient use of hardware resources. 

The multiple CLPs design involves optimizing the number of  CLPs used, assignment of CONV layers to CLPs, the parallelization factors $\langle T_{n}, T_{m} \rangle$ for each CLP, and the tilling parameters $\langle T_{r}, T_{c} \rangle$ for each CONV layer. This imposes an intractable design space where it is almost impossible to find near-optimal configurations through exhaustive search algorithms. Therefore, in this work, we employed SA and TS algorithms to design CNN accelerators using multiple CLPs on FPGAs targeting the optimization of the number of execution cycles. 

Experimental results on accelerating AlexNet, VGGNet, SqueezeNet~$1.1$, and GoogLeNet architectures demonstrated the effectiveness of the proposed optimization framework in providing efficient accelerators. The results also show the importance of using metaheuristics in finding different configurations
for a given CNN using a systematic methodology. Compared to a single CLP design, $\textnormal{SA-based}$ $\textnormal{Multi-CLP}$ and $\textnormal{TS-based}$ $\textnormal{Multi-CLP}$ accelerate CNN computations by about $1.31\times$ $-$ $2.37\times$ when targeting Xilinx Virtex-7 FPGAs. Our implementation achieves a performance of $113.92$ GFLOPs under $100$ MHz working frequency.

In future work, we consider improving the quality of resource usage model in predicting DSP slices required by taking into account those DSPs used in control logic, loop indexing, address calculation, etc. Additionally, we target the adoption of data quantization technique for further acceleration. Reducing the bit-precision level for features and weights in turn allows multipliers and accumulators to be implemented using other logical elements not only using DSPs. This increase in computing units enhances CNN throughput further. 

\section{Acknowledgment}

The authors would like to thank King Fahd University of Petroleum and Minerals (KFUPM) for supporting this research and providing the computing facilities.

\printbibliography

\end{document}